\documentclass[journal]{IEEEtran}
\usepackage{color}
\usepackage{float}
\usepackage{amssymb}
\usepackage{multirow}
\usepackage{graphicx} 
\usepackage{cite}
\usepackage{amsfonts}
\usepackage{booktabs}
\usepackage{subcaption}
\usepackage{amsmath}
\usepackage{array}
\usepackage{hyperref}
\usepackage{soul}
\soulregister\cite7 
\soulregister\citep7 
\soulregister\citet7 
\soulregister\ref7 
\soulregister\pageref7 

\ifCLASSINFOpdf
\else
\fi
\begin{document}
%
\title{Skeleton-based Action Recognition through Contrasting Two-Stream Spatial-Temporal Networks }

\author{Chen Pang, Xuequan Lu, Lei Lyu$^*$
\thanks{Chen Pang and Lei Lyu are with School of Information Science and Engineering, Shandong Normal University, Jinan 250358, China.}
\thanks{Xuequan Lu is with School of Information Technology, Deakin University, Geelong, Australia.}
\thanks{$^*$Corresponding author: Lei Lyu (e-mail: lvlei@sdnu.edu.cn) }
}

\maketitle

\begin{abstract}

For pursuing accurate skeleton-based action recognition, most prior methods use the strategy of combining Graph Convolution Networks (GCNs) with attention-based methods in a serial way. However, they regard the human skeleton as a complete graph, resulting in less variations between different actions (e.g., the connection between the elbow and head in action ``clapping hands'').  
For this, we propose a novel Contrastive GCN-Transformer Network (ConGT) which fuses the spatial and temporal modules in a parallel way. The ConGT involves two parallel streams: Spatial-Temporal Graph Convolution stream (STG) and Spatial-Temporal Transformer stream (STT). The STG is designed to obtain action representations maintaining the natural topology structure of the human skeleton. The STT is devised to acquire action representations containing the global relationships among joints. Since the action representations produced from these two streams contain different characteristics, and each of them knows little information of the other, we introduce the contrastive learning paradigm to guide their output representations of the same sample to be as close as possible in a self-supervised manner. Through the contrastive learning, they can learn information from each other to enrich the action features by maximizing the mutual information between the two types of action representations. To further improve action recognition accuracy, we introduce the Cyclical Focal Loss (CFL) which can focus on confident training samples in early training epochs, with an increasing focus on hard samples during the middle epochs.
We conduct experiments on three benchmark datasets, which demonstrate that our model achieves state-of-the-art performance in action recognition.

\end{abstract}


%
\IEEEpeerreviewmaketitle

\section{Introduction}
%
%
%
%
\IEEEPARstart{H}{uman} action recognition has become a fundamental task in computer vision which is extensively applied in many real-world applications, such as intelligent security \cite{6}, virtual reality \cite{7}, and human–machine interaction \cite{8}. Skeleton-based action recognition task has received significant attention due to its computation efficiency and robustness against viewpoints or appearance. 

The core of skeleton-based action recognition is to learn the discriminative representations of skeleton sequences. At present, many deep learning based methods have achieved excellent performance by using Convolutional Neural Networks (CNNs) \cite{24,25}, Recurrent Neural Networks (RNNs) \cite{29,31} to learn the action representations based on the specific recognition task. However, these methods rarely consider the co-dependency contained in body joints and ignore some important motion information. To better capture joint dependencies, Graph Convolutional Networks (GCNs) are exploited to aggregate information based on body structures. Spatio-Temporal Graph Convolutional Network (ST-GCN) \cite{13} is a pioneering work to model the skeleton data as a spatio-temporal graph with the joints as graph nodes and natural connections in both human body structures and time as graph edges. Later, many variants  \cite{14,39,40} are extended based on ST-GCN, following the same strategy. Although GCNs have been proved to perform well on skeleton data, they still have some limitations. First, in GCN-based methods, the human body is represented as a predefined graph fixed over all actions, ignoring certain implicit relations between nonadjacent joints, such as the connection between hand and head during touching head. Second, with the deepening of graph convolutional layers, the probability of over-smoothing problem will increase that the representations of neighbor nodes tend to converge to the same value, which causes confusion between joints. Third, in most existing GCN-based methods, the temporal connections between remote frames are underestimated since the temporal convolution operations are limited in a local neighborhood. 
To cope with these defects, researchers introduced attention modules behind the GCN layers to effectively capture the long-distance relations in the supervised manner. Shi et al. \cite{48} designed a decoupled spatial-temporal attention network to calculate the connections between each pair of joints without knowing their positions or mutual connections. In \cite{17}, ST-TR used transformer to capture the relations of each pair of nodes, ignoring the inherent topology of the human skeleton. Although the recognition accuracy can be improved by combining the GCN layers with attention modules in a serial manner, each node is treated in isolation and the human skeleton is regarded as a complete graph with connections built between each joint and the rest joints, resulting in less variations between different actions. Taking the two actions of ``clapping hands'' and ``touching nose'' for example, the connections among the most joints are considered to be same, except for the stronger connection between the hands in ``clapping hands'' and the stronger dependence between the hands and nose in ``touching nose''. While the connection between the elbow and head should not be considered in ``clapping hands'', it is helpful in ``touching nose''. Therefore, it is essential to capture the long-distance relations for better action recognition while preserving the primitive human skeleton structure. 

To this end, we propose a novel Contrastive GCN-Transformer Network (ConGT) that considers GCN and attention model in a parallel manner. Specifically, the network contains two parallel streams, i. e. Spatial-Temporal Graph Convolution stream (STG) and Spatial-Temporal Transformer stream (STT). The STG is used to extract the joint relationships based on the topology of the human skeleton graph, consisting of adaptive GCN module (AGCN) and temporal convolutional network module (TCN). In particular, the AGCN is designed to enforce the generated graph to reflect the relationships of joints flexibly. Different from the STG, the STT is primarily responsible for accurately capturing the relationships among arbitrary joints in the intra- and inter- frames, which is comprised of spatial transformer module and temporal transformer module. Since the action representations produced from these two streams contain different characteristics and each of them knows little information of the other, we introduce the contrastive learning paradigm to guide their output representations of the same sample to be as close as possible in the embedding space in a self-supervised manner. Specifically, the action representations learned by STG involve the natural topology of the human skeleton and the action representations learned by STT involve long-distance relations. Through the contrastive learning paradigm, they can learn information from each other to enrich the action features by maximizing the mutual information between the two types of action representations. Moreover, to further improve the performance of our model, we introduce Cyclical Focal Loss (CFL) instead of Cross-Entropy loss as our learning objective. In contrast to the Cross-Entropy loss, the CFL can focus on confident training samples in early training epochs of our model, with an increasing focus on hard samples during the middle epochs.

In summary, the main contributions of our work can be concluded as follows: 
\begin{itemize}
\item[$\bullet$] We propose a novel Contrastive GCN-Transformer Network (ConGT), which can capture the relationships between arbitrary joints in intra- and inter- frames more accurately while maintaining the topology structure of human skeleton graph. 

\item[$\bullet$] We propose a Spatial-Temporal Graph Convolution stream (STG) with an adaptive graph strategy and a Spatial-Temporal Transformer stream (STT) to learn the action representations containing local and global joints relations.

\item[$\bullet$] We introduce the contrastive learning paradigm to integrate the information of two types of action representations by maximizing mutual information between them. 

\item[$\bullet$] We introduce the Cyclical Focal Loss (CFL) as the learning objective of our network to improve the action accuracy. 
\end{itemize}

\section{Related work}
\subsection{Action Recognition}
In this section, we will briefly review the related works about the three fields on skeleton-based action recognition: Convolutional Neural Networks (CNNs) based methods, Recurrent Neural Networks (RNNs) based methods and Graph Convolutional Networks (GCNs) based methods. 

\subsubsection{CNN-based Methods}
In the early work, the mainstream network is based on CNN and RNN. CNN is mainly used to process 2D images, which can easily learn the high-level semantic features of images. Thus, most CNN-based methods generally encode skeleton features to 2D pseudo images. In \cite{24}, each skeleton sequence was transformed into three clips which were correspond to every channel of the cylindrical coordinates of the skeleton sequence. Then the frames of the clips are jointly processed by a multi-task learning network. Wang et al. \cite{25} encoded the spatio-temporal information carried in skeleton sequence as joint trajectory maps. Li et al. \cite{23} proposed an end-to-end hierarchical co-occurrence network to learn the con-occurrence feature with a hierarchical methodology, where different levels of contextual information is aggregated gradually. They used the 3D coordinates of joints to generate several 2D images that are sent into the pretrained VGG-19 for action recognition. Although the spatial features can be reserved in these pseudo images, the motion information contained in actions is ignored. To solve this problem, Rong et al. \cite{26} used geometric algebra to learn the shape-motion representations and applied the multi-stream CNN models to fuse the complementary shape-motion representations. But 2D CNNs have a weak ability to capture the temporal and spatial features in the human skeleton. Later, Duan et al. \cite{27} proposed C3D network instead of 2D CNNs, which uses a heat map to denote the spatial correlations of joints and uses a stack to present the temporal sequence. Although this approach has yielded good results, it still ignores the motion correlation of the skeleton data which is a common challenge of CNN-based methods.

\subsubsection{RNN-based Methods}
Recurrent neural networks (RNNs) \cite{28} can process sequence data with variable lengths because its cellular states can determine which temporal states should be left and which should be forgotten. Therefore, it has more advantages in processing temporal sequences. In the field of action recognition, RNN-based methods represent the skeleton data as a vector sequence that contains the location information of all joints in one frame. Du et al. \cite{29} divided the human skeleton into five parts and fed them to five hierarchical RNN networks separately. In \cite{31}, Liu et al. introduced a new gating mechanism into Long Short Term Memory(LSTM) network to handle the noise and occlusion in 3D skeleton data. Lee et al. \cite{30} proposed ensemble Temporal Sliding LSTM (TS-LSTM) networks containing short-term, medium-term and long-term TS-LSTM. They focused on the temporal correlation of various human body parts but ignored the spatial structure of the skeleton. In most of the actions, the variation range of action in the space domain is larger than that in the time domain. So, researchers have also been trying to design some RNN-based networks to process spatial information. For example, in \cite{312}, Liu et al. proposed a global-aware attention LSTM to make use of the global contextual information, which can selectively focus on the informative joints in each frame of the skeleton sequence and further enhance attention to spatial information. However, how to perceive the spatial correlations of the human skeleton remains a burning challenge for RNNs.

\subsubsection{GCN-based Methods}
Because the topology of skeleton data is encoded in the form of graphs rather than two-dimensional grids or vector sequences, CNN-based methods or RNN-based methods may not be the optimal choice. Recently, Graph Convolution Networks (GCNs) have achieved remarkable results in many works based on graph structure data, which can be divided into two types: spatial GCNs \cite{31,32,33} and spectral GCNs \cite{34,35,36}. For spectral GCNs, the input graphs are first transformed into the spectral domains and then operated by means of Fourier transform. Spatial GCNs are applied directly to the nodes of the graph and their neighbors, which are more similar to the traditional convolution neural networks. Our work follows the spatial GCN methods. 

Spatial-Temporal Graph Convolutional Networks (ST-GCN) \cite{13} is the pioneering method to model the skeleton data, which breaks the limitations that the previous method cannot effectively extract spatial and temporal features at the same time. ST-GCN models the joint connection and extracts correlated features as a spatio-temporal graph, where the graph convolution operates on the spatial features, and the 2D convolution operates on the temporal motion correlations. Recently, many works also adopt the same strategy. Li et al. \cite{39} combined the actional links and structural links into a generalized skeleton graph and used the actional-structural graph convolution and temporal convolution to learn the spatial-temporal features. In \cite{40}, Li et al. proposed a spatio-temporal graph routing scheme to adaptively learn the high-order connectivity relationships for physically-apart skeleton joints.  Specifically, spatial graph routing aims at spatial relationships based on sub-group clustering, while the temporal graph routing explores the temporal correlations. Shi et al. \cite{14} proposed a two-stream adaptive graph convolutional network to make the value of the adjacency matrix be variable. With the adaptive strategy and the two-stream pattern, this method can model both human joints features and human bones features simultaneously. DGNN \cite{41} leveraged an alternating spatial aggregation scheme to update the joint and bone features. Liu et al. \cite{42} proposed a disentangling and unifying graph convolutional network, including a simple disentangled multi-scale graph convolution and G3D module. The former is used to disentangle the importance of nodes in different neighborhoods, which can model the long-range relationships. The latter is presented to directly propagate the information across the spatial-temporal graph by leveraging the cross-spacetime edges as skip connections.

\subsection{Transformer in Computer Vision}
Transformer \cite{43} was proposed for the Natural Language Processing tasks to make up for the shortcoming of the RNN methods. The great contribution of transformer is the self-attention, which can dynamically focus on the global context information. Alexey et al. \cite{44} applied a pure vision transformer to sequences of image patches, which has achieved excellent performance on the image classification task. In the object detection field, Carion et al. \cite{45} proposed detection transformer reasoning about the relations of the objects and the global image context. Wang et al. \cite{46} proposed Max-DeepLab for semantic segmentation, which directly predicts class-labelled mask with a mask transformer. Zhou et al. \cite{47} proposed a masked transformer for video understanding tasks. For skeleton-based action recognition, Shi et al. \cite{48} proposed a pure transformer network to model the correlation between joints without using the traditional skeleton graph representation. They designed a decoupled spatial-temporal attention network to calculate the attention score between each pair of joints without knowing their positions or mutual connections. Similarly, Plizzari et al. \cite{17} proposed a spatial and temporal transformer network. The spatial self-attention module is used to capture the intra-frame correlations between human joints, while the temporal self-attention module is used to model the inter-frame relationships. However, these methods have a common problem that they ignore the inherent topology of the human skeleton and overestimate the correlation between some joints. In this way, the relationship that does not exist between some joints in some actions will also be forced to be imposed through the calculation of attention score. For example, in the action of ``sitting down'', it is not necessary to capture the relationship between the left hand and the right hand that will bring trouble for the model to recognize actions. In our work, we use the transformer structure to enhance the ability of the GCN to capture relationships between joints. Different from the original transformer and its variants, the position encoding is not used in our work because of the topological invariance of graphs. 

\begin{figure*}[h]
	\centering  
	\includegraphics [scale=0.5]{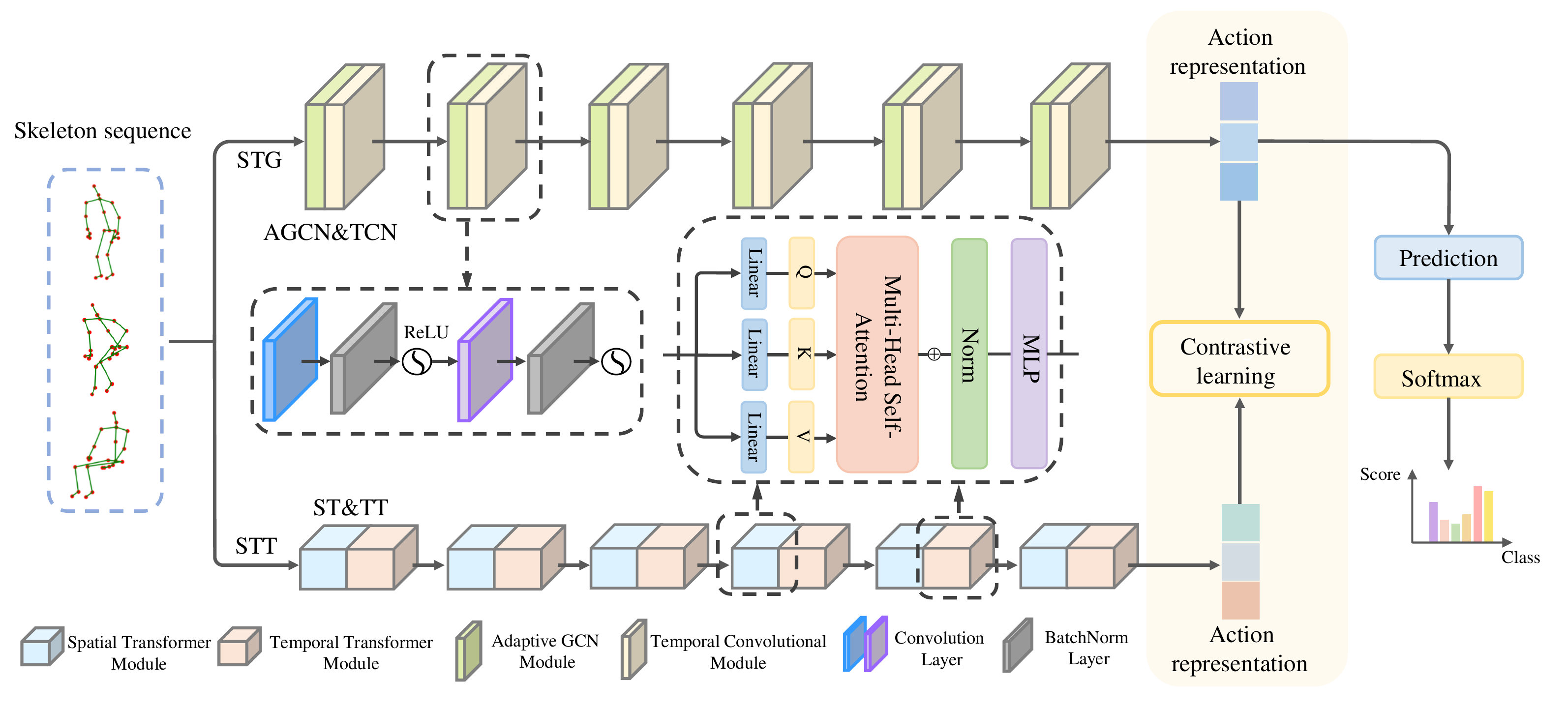}
	
	\centering  \caption  {The overall architecture of the proposed ConGT. The skeleton sequence is first fed into two streams, where the spatial-temporal GCN stream (STG) processes the input graph through the adaptive GCN  module (AGCN) and temporal convolutional module (TCN). The spatial-temporal transformer stream (STT) operates on the input graph with spatial transformer (ST) and temporal transformer (TT) modules. The ST and TT modules have the same structure. Then the contrastive learning maximizes the mutual information across the two streams. Finally, the classifier is added for action classification. And in the test phase, we choose the prediction result of the output feature of STG as the final result of the network.}
	\label{framework}
\end{figure*}

\subsection{Self-Supervised Learning}
The intention of self-supervised learning (SSL) is to learn the internal structures of data and the feature representations from the unlabeled data. In \cite{50}, it has been verified that self-supervised pre-training also can bring some assistance for supervised learning. SSL was first used in visual representation \cite{51}, and now it has a number of applications in computer vision fields \cite{52,53},\cite{54}. It is usually achieved by pretext task which is a hot topic of research. In \cite{58}, they predicted the arrangement of multiple shuffled image patches by using SSL to learn the spatial relationships. Chen et al. \cite{59} proposed a simple framework for contrastive learning of visual representations. They 
force the feature representation between positive samples to be more similar than those between negative ones. For the sequential data, some methods \cite{57,60} learn the temporal features by predicting the sequential order of sampled frames or clips. Cho et al. \cite{61} proposed a video representation method via a prediction task. For skeleton-based tasks, Lin et al. \cite{77} integrated prediction task, recognition task, and contrastive learning paradigm to learn skeleton features from different aspects. Zheng et al. \cite{75} explored an unsupervised representation learning approach to compactly encode long-term global motion dynamics. Su et al. \cite{76} proposed an unsupervised encoder-decoder recurrent neural network to cluster similar movements. Xu et al. \cite{78} proposed an unsupervised framework based on encoder-decoder structure to extract more discriminative temporal features and explored the inherent action similarity within the action encoding by clustering. Rao et al. \cite{79} utilized a variety of data enhancement strategies on unlabeled data to obtain the action representations with the contrastive learning paradigm. Li et al. \cite{80} proposed a cross-view contrastive learning model by leveraging multi-view complementary supervision signal. Wang et al. \cite{70} proposed the contrast-reconstruction representation learning network to capture postures and motion dynamics simultaneously. In \cite{73}, Guo et al. utilized the abundant information mining strategy to make better use of the movement patterns. In \cite{55,56}, it is suggested that contrasting congruent and incongruent views of graphs with mutual information maximization can help encode rich representations. Inspired by them, we also integrate contrastive learning into the training of our network to enhance graph modelling by maximizing mutual information between two types of action representations and improve the accuracy of action recognition.

\section{Method}
In this section, we first present an overview of Contrastive GCN-Transformer Network (ConGT) and describe the details of each component of the ConGT. Then, we show the process of enhancing ConGT with contrastive learning. Finally, we depict the details of Cyclical Focal Loss (CFL).

\subsection{ConGT}
Our proposed ConGT includes two streams: Spatial-Temporal Graph Convolution stream (STG) and Spatial-Temporal Transformer stream (STT), as illustrated in Fig. \ref{framework}. Specifically, STG is used to obtain the action features based on the topology of the human skeleton graph,  consisting of adaptive graph convolutional network module (AGCN) and temporal convolutional network module (TCN). The AGCN learns the topology of the graph for different layers and skeleton samples while the TCN models the temporal connections between adjacent frames in time dimension. Likewise, the STT contains spatial transformer module and temporal transformer module, which are primarily responsible for accurately capturing the relationships between arbitrary joints in the intra- and inter- frames. Both streams can output action representations, but the generated action representations have different characteristics and each knows little information of the other. For this reason, we introduce the contrastive learning paradigm into ConGT to enforce the two streams learn more distinctive information. Through the contrastive learning, we can maximize the interactive information from the representations learned by these two streams. In this end, we can obtain the predicted class for the input skeleton graph with the classifier. 

\begin{figure*}[h]
	\centering  
	\includegraphics [scale=0.7]{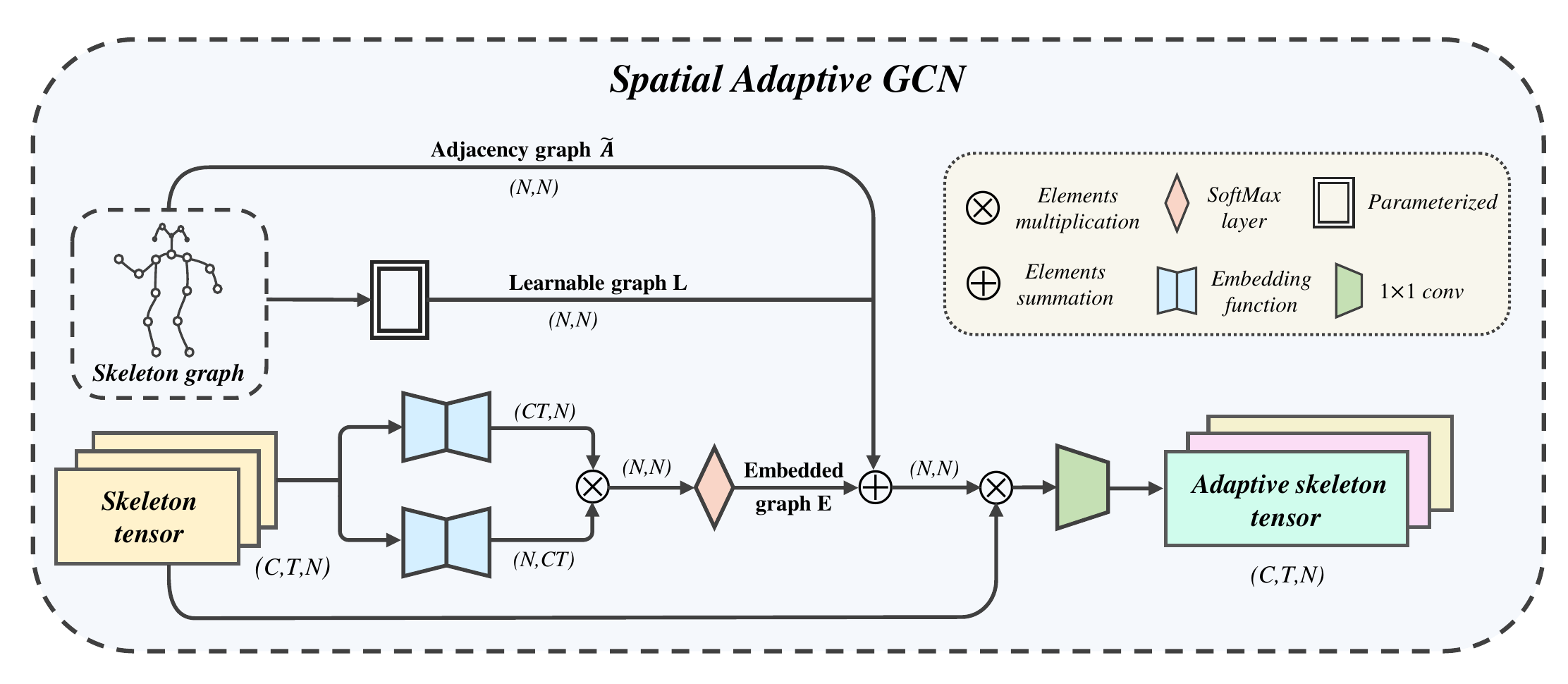}
	
	\centering  \caption  {The schematic diagram of the spatial adaptive graph convolution. The input consists of an adjacency matrix of the skeleton graph and a skeleton tensor. The learnable matrix is a parameterized matrix and is trained with the whole model, and the skeleton tensor passes through the embedding function to obtain the embedded matrix. Then the three types of matrices are added together and multiplied with the skeleton tensor to get the output adaptive skeleton tensor.}
	\label{AGCN}
\end{figure*}

\subsection{Spatial-Temporal Graph Convolution Stream}
\textbf{Notation.} For skeleton-based action recognition, the human action is represented as a sequence of skeleton frames. In a frame, each skeleton is expressed as a graph $G=(V,E)$, in which $V=\{v_i|(i=1,\ldots\,N)\}$ denotes the collection of vertices representing $N$ human joints and $E=\{e_{i,j}|(v_i,\ v_j)\}$ denotes the collection of edges representing the human bones. Formally, the adjacent matrix $A\in R^{N\times N}$ is used to describe the structure of skeleton graph, where the value of $A_{ij}$ is 0 or 1 indicating whether an edge exists between joints $v_i$ and $v_j$. And the feature tensor $X\in R^{C\times T\times N}$ denotes the sequence of skeleton frames, where $C$ represents the coordinate dimension, $T$ denotes the total number of skeleton frames contained in the sequence, and $N$ is the total number of human joints. 

\textbf{Graph Convolutional Network.} The process of updating a graph by GCN is to aggregate nodes information with edge information to generate a new graph representation. For skeleton-based action recognition, the action representations can be obtained through GCN operating on the adjacent matrix ${A}$. Let $H$ denote the action representation which is initialized to ${H^{(0)}=X}_{in}=\left\{X_{in}^1,X_{in}^2,\ \ldots,X_{in}^n\right\}\in R^{C\times T\times N}$, where $X_{in}$ is the graph representation of input skeleton frame, $T$ and $N$ are the total number of the frames and human joints, respectively. Then the graph convolution operation can be expressed as: 
\begin{equation}
\centering
H_t^{\left(l+1\right)}=D^{-\frac{1}{2}}\bar{A}D^\frac{1}{2}H_t^l\omega^l
\label{eq1}
\end{equation}
where $\bar{A}=A+I$, $I$ is the identity matrix, $D$ is the degree matrix of $\bar{A}$, $H_t^l$ denotes the skeleton tensor in the $t$-th frame at the $l$-th layer, and $\omega^l\in R^{C^l\times C^{(l+1)}}$ is a learnable weight matrix. 

In the implementation of GCN, the higher-order polynomial of the adjacency matrix $A$ is applied to aggregate the skeleton tensor $H$ to get the high-order neighbor information. Thus, Eq. \ref{eq1} can be rewritten as: 
\begin{equation}
\centering
H_{t}^{(l+1)}=\sigma\left(\sum_{k=0}^{K_{v}}\left(\tilde{A}^{k} H_{t}^{l}\right)\right) \odot \omega_{k}^{l}
\label{eq2}
\end{equation}
where $K_v$ denotes kernel size in the spatial dimension. $\tilde{A}$ represents the normalized adjacency matrix, while ${\widetilde{A}}^k$ represents $k$-power of the normalized adjacency matrix which can represent the relationship between adjacent nodes of order $k$. $\omega_k^l\in R^{C^l\times C^{(l+1)}}$ is a learnable weight matrix, $\odot$ denotes the dot product, and $\sigma(\cdot)$ denotes the activation function.

\textbf{Adaptive graph strategy.} In the traditional graph convolution, a common way to calculate graph topological relationships is using the adjacency matrix. Each element of the adjacency matrix $A$ has a value of 0 or 1. When $A(i,j)=0$, it means there is no connection between the joints $v_i$ and $v_j$, otherwise adjacent. Note that even if after several multiplication operations between the adjacency matrix and skeleton tensor, $A(i,j)=0$ will still exist, indicating that there is no relationship between joints $v_i$ and $v_j$ during the model training. This will ignore the connection between some joints and further affect the recognition accuracy. For instance, the two hands in action ``clapping hands" are not directly linked, but the interaction between them is actually very useful for recognizing this action. To this end, we employ the adaptive graph strategy \cite{14} to adaptively reflect the relationships of joints.

As shown in Fig. \ref{AGCN}, the adaptive graph strategy is implemented by adding the adjacency matrix, the learnable matrix, and the embedded matrix together. Then the graph convolution integrated with the adaptive graph strategy can be formulated as: 
\begin{equation}
\centering
H_t^{\left(l+1\right)}=\sigma\left(\sum_{k=0}^{K_v}\left(\left(\widetilde{A}+L+E\right)^kH_t^l\right)\right)\odot\omega_k^l
\end{equation}
where the normalized adjacency matrix $\widetilde{A}\in R^{N\times N}$ denotes the original topology of the skeleton graph. $L\in R^{N\times N}$ is a learnable matrix which is initialized with the adjacent matrix $A$ to accelerate the convergence of the network. As we perform actions, our joints move in groups, but the importance of each joint is different in different groups. Therefore, we need to define an importance weight to scale the contribution of node features to neighboring nodes. The learned matrix $L$ is an attention map that indicates the importance of each node. In this way, the data-driven dependencies between nonadjacent joints $v_i$ and $v_j$ can be generated as the depth of the network increases. The embedded matrix $E\in R^{N\times N}$ learns an individual graph for each sample. In these individual graphs, the weights of edges are calculated by measuring the similarity of graph nodes which can be obtained by the normalized embedded Gaussian function. The elements of the embedded matrix can represent the strength of the dependency between two joints, and the value of them ranges from 0 to 1. The embedded matrix $E$ is described mathematically as follows: 
\begin{figure*}[h]
	\centering  
	\includegraphics [scale=0.6]{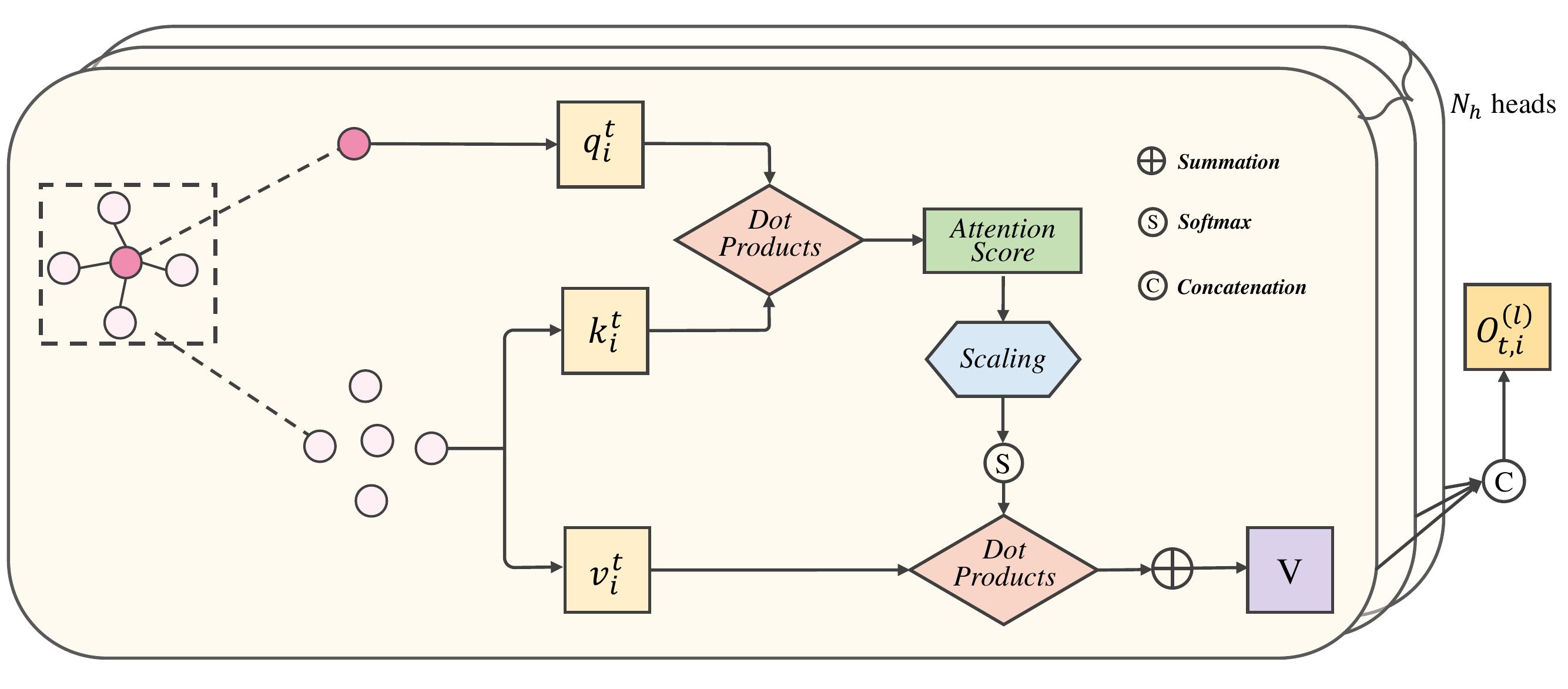}
	
	\centering	\caption{The detail of  multi-head attention. } 
	\label{multi-attention}
\end{figure*}
\begin{equation}
\centering
E=softmax\left(\left(\theta_{1}\left(f^{(l)}\right) \cdot \delta_{1}^{(l)}\right)^{T} \cdot\left(\theta_{2}\left(f^{(l)}\right) \cdot \delta_{2}^{(l)}\right)\right)
\end{equation}
where $\theta$ represents the embedding function that can map any two joint vectors to the same vector space. $\delta$ denotes the parameters of the embedding function. $f^{\left(l\right)}\in R^{C\times T\times N}$ denotes the feature tensor of the $l$-th layer. And it will be converted into two intermediate embeddings by embedding functions $\theta_1$ and $\theta_2$. Then the two embeddings are multiplied to get the embedded graph $E$.

\textbf{Temporal convolution Network.}  Unlike the spatial topology, the temporal topology of joints is linear. Thus, temporal relationships are usually captured by using ordinary convolution operation rather than graph convolution.

In general, the spatial-temporal convolution on the skeleton graph can be formulated as:
\begin{equation}
\centering
H^{\left(l+1\right)}=TCN\left(AGCN\left(H^{\left(l\right)}\right)\right)
\end{equation}
where $TCN(\cdot)$ is a temporal convolution with a kernel size $K_T\times1$, and $K_T=9$ in this work. $AGCN(\cdot)$ denotes the adaptive graph convolution network, and $H^{\left(l\right)}$ denotes the action representation in the $l$-th layer.

\subsection{Spatial-Temporal Transformer Stream}
Although the adaptive graph strategy can make up for the defect that $A(i,j)=0$ cannot be replaced in multiplication, the long-distance connections are still easily underestimated. Moreover, with the stacking of layers, the risk of over-smoothing of graph convolution will increase, leading to poor ability to accurately capture the relationships between joints which are depicted by the action representations. For these reasons, we propose the spatial-temporal transformer stream (STT) to enhance the dependence of local neighboring joints and further  effectively capture long-distance relationships in both spatial and temporal dimensions. 

The work flow of the spatial-temporal transformer is depicted in Fig. \ref{framework}. In space, for each node $n_i^t$ of the skeleton graph in the frame $t$, a query vector $q_i^t$, a key vector $k_i^t$, and a value vector $v_i^t$ can be calculated by the trainable linear transformations with three parameters matrices $W_q\in R^{C_{in}\times d_q}$, $W_k\in R^{C_{in}\times d_k}$, and $W_v\in R^{C_{in}\times d_v}$, which are shared by all nodes. With these vectors, we use the multi-head attention to calculate the attention score to weight the value vector $v_i^t$ that corresponds to the query vectors. In time dimension, a node in one frame will pay attention to nodes representing the same joint in other frames.  
Similar to the spatial transformer, in temporal transformer, the first step is also to calculate a query vector $q_{fj}$, a key vector $k_{fj}$, and a value vector $v_{fj}$ for each node $n_{fj}$ in different frame $f$. Then, we compute the attention score with multi-head attention, which is used to determine how much attention one node is paid to other nodes that represent the same joints along the temporal dimension.

As the core component of transformer, the multi-head attention is detailed in Fig.  \ref{multi-attention}. The input consists of the query vector $q$, the key vector $k$, and the value vector $v$. Then, the dot products of the query with all keys are calculated to get the attention scores between all nodes. In order to prevent the vanishing gradient in the process of backpropagation, we scale the attention scores by $\frac{1}{\sqrt{d_k}}$, where $d_k$ denotes the dimension of query vector and key vector. Subsequently, we apply the softmax function on the attention scores to weight the value vectors. Finally, the attention-enhanced value vectors of nodes are aggregated in a summation manner. 
The above process is repeated $N_h$ times with different queries, keys and values, where $N_h=8$ in our work. Then the results of the $N_h$ attention heads are concatenated together to constitute the output representation. Formally, this process is expressed as: 
\begin{equation}
\centering
O_{i}^{(l)}=\prod_{1}^{N_{h}}\left(\sum_{j \in N_{i}} softmax\left(\frac{S_{i, j}^{(l)}}{\sqrt{d_{k}}}\right) v_{j}^{(l)}\right)
\end{equation} 
\begin{equation}
\centering
S_{i, j}^{(l)}=q_{i}^{(l)} \cdot k_{j}^{(l)}
\end{equation}
where $O_{i}^{\left(l\right)}$ denotes the output of the multi-head attention.  $\prod_{1}^{N_h}(\cdot)$ denotes the concatenation of $N_h$ heads. $q_{i}^{(l)}$, $k_{i}^{(l)}$, and $v_{i}^{(l)}$ denote the query vector, the key vector, and the value vector of node $i$ in the frame $t$ at the $l$-th layer, respectively. $S_{i,j}^{\left(l\right)}$ is the attention score between nodes $i$ and $j$, which is treated as a weight when aggregating the values of different nodes. 

Next, the output of the multi-head attention passes through a batch normalization layer and a Multi-Layer Perceptron (MLP) block to obtain the input tensor of the next layer or the final output of STT in the last layer. The STT can be formulated as:
\begin{equation}
\centering
H^{\left(l+1\right)}=TT\left(ST\left(H^{\left(l\right)}\right)\right)
\end{equation}
where $TT(\cdot)$ and $ST(\cdot)$ denote the temporal transformer module and the spatial transformer module, respectively. $H^{(l)}$ denotes the generated action representation in the $l$-th transformer layer.

\subsection{Enhance ConGT with Contrastive Learning}
The STG can learn the topology of the graph for different GCN layers and skeleton samples in an end-to-end manner, obtaining the action representations based on the topology of the human skeleton graph. However, the implicit relations between nonadjacent joints are ignored, such as the connection between hand and head during touching head. Meanwhile, the temporal connection between remote frames is underestimated due to the limitation of temporal convolution kernel size. 
To solve them, we bridge GCN and attention module in a parallel way with contrastive learning to enrich the action features. We design the STT based on transformer to capture the long-distance correlations between each pair of joints in both spatial and temporal dimensions without knowing their positions or mutual connections. Since each stream encodes a representation that only depicts either the topology of human skeleton graph or the long-distance correlations between each pair of joints, the two types of action representations know little about each other but may mutually complement each other. Specifically, they can be the ground-truth of each other for contrastive learning. Then, by maximizing the mutual information between the action representations learned via the two streams through contrastive learning, our network can learn more distinctive information to enhance recognition accuracy.  

We adopt InfoNCE as the contrastive learning objective, which is defined as:
 
\begin{equation}
\centering
 L_{con }=-\log \sigma\left(f_{d}\left(a_{i}^{g}, a_{i}^{t}\right)\right)-\log \sigma\left(1-f_{d}\left(\tilde{a}_{i}^{g}, a_{i}^{t}\right)\right)
\end{equation}
where $a_i^g$ and $a_i^t$ denote the action representations obtained through the STG and STT, respectively. ${\widetilde{a}}_i^g$ represents the negative samples obtained by corrupting $a_i^g$ with both row-wise and column-wise shuffling. $f_d(\cdot)$ is a discriminant function: $R^d\times R^d\rightarrow R$, which scores the agreement between the two input vectors. The contrastive learning objective is to maximize the agreement of two different types of  representations, while minimizing the agreement with other negative representations. In this way, both STG and STT can acquire information from each other and further enrich the action features.

In the end, we combine the contrastive learning task with the action recognition task to form multi-task learning, where contrastive learning is the auxiliary task. We optimize our model by means of joint learning that is defined as: 
\begin{equation}
\centering
L=L_{CFL}+\beta L_{con}
\end{equation}
where $L_{CFL}$ is cyclical focal loss that is the learning objective of action recognition task and $\beta$ is a hyper-parameter used to control the magnitude of the contrastive task.

\subsection{Cyclical Focal Loss}
We define the learning objective of action recognition task as the cyclical focal loss which is formed of the focal loss and the general cyclical training principle. Cyclical focal loss can focus on confident predictions in the early epochs of the network training. As the number of training epochs increases, it will focus more on misclassified samples. In the following, we will describe the details of the cyclical focal loss. 
 
Focal Loss is mostly used for binary classification problems. It modifies cross-entropy softmax loss to reduce the weight of easily classified samples so that the model can focus more on samples that are difficult to classify during training. It is defined as:
\begin{equation}
\centering
L_{l c}=-(1-p_t)^\gamma log(p_t)
\label{eq11}
\end{equation}
where $p_t$ denotes the softmax probabilities, and $\gamma\geq0$ is a tunable hyper-parameter. For the confident training samples, the value of $p_t$ tends to be 1, and the weight  $\left(1-p_t\right)^\gamma$ for the loss will drive the loss to zero faster than that for cross-entropy. Although the focus loss is beneficial in tasks with unbalanced class data, it often affects performance when the dataset is more balanced. Therefore, in most applications, the focus loss is not the best choice. 

Combining the focus loss with the general cyclical training principle, the cyclical focal loss \cite{64} is proposed. In \cite{63}, the general cyclical training of a neural network is considered as a combination of curriculum learning in the early epochs with fine-tuning at the end of training. Specifically, the easy and confident training samples are used at the start and end stages of network training, and the hard training samples are processed at the middle stage of training. For this purpose, Smith et al. \cite{64} proposed a new loss: 
\begin{equation}
\centering
L_{h c}=-\left(1+p_{t}\right)^{\gamma_{h c}} \log \left(p_{t}\right)
\label{eq12}
\end{equation}
where $p_t$  denotes the softmax probabilities, and $\gamma_{hc}\geq0$ is a tunable hyper-parameter. In this manner, the loss can pay more attention to the confident training samples. And the hard training samples can be heavily weighted via Eq. \ref{eq11}. Therefore, the cyclical focal loss can be accomplished by combining Eq. \ref{eq11} and Eq. \ref{eq12} in a reasonable manner. In \cite{64}, they used a linear schedule and defined a parameter $\xi$ to combine them that varies with the training epoch as: 
\begin{equation}
\centering
\xi= 
\begin{cases}
	1-f_{c} \frac{ep_{i}}{ep_{n}} & \text { if } f_{c} \times ep_{i} \leq ep_{n} \\
	{\left(f_{c} \frac{ep_{i}}{ep_{n}}-1\right) /\left(f_{c}-1\right)} & \text { otherwise }
\end{cases}
\label{eq13}
\end{equation}
where $ep_i$ denotes the number of current training epochs and $ep_n$ is the total number of training epochs. 
$f_c$ denotes the cyclical factor that provides adaptability for the cyclical schedule. 

Integrating Eq. \ref{eq11}, Eq. \ref{eq12} and Eq. \ref{eq13}, the cyclical focal loss can be defined as:
\begin{equation}
\centering
C F L(p, y)=\xi L_{h c}+(1-\xi) L_{l c}
\end{equation}

In our experiments, we keep the values of the hyper-parameters consistent with those in the original work, where $\gamma_{lc}=2$,  $\gamma_{hc}=2$, $f_c=4$.

\section{Experiments}
To verify the effect of the proposed method, we conduct extensive experiments on three widely used datasets: NTU-RGBD 60 \cite{65}, NTU-RGBD 120 \cite{66}, and Northwestern-UCLA \cite{81}. In this section, we first give the description of the three datasets in detail. Next, we will describe the experiment settings. Then, the comparisons between the proposed method and several state-of-the-art methods are introduced. Finally, we investigate the contributions of each component in the proposed method.

\subsection{Datasets}
\textbf{NTU-RGBD 60 (NTU-60):} NTU-60 is one of the widely used datasets for skeleton-based action recognition tasks, which contains 56,880 samples of 60 different classes. The dataset is captured by a Microsoft Kinect V2 camera. The skeleton data is formed of the 3D joint locations (X, Y, Z) of 25 joints. In each video, there are no more than two persons. The dataset is divided into Cross-View (X-View) Setting and Cross-Subject (X-Sub) Setting. In X-View, the actions are captured by three cameras with the same height in the vertical direction and different angles $-45^{\circ}, 0^{\circ}, 45^{\circ}$ 
in the horizontal direction, where the training set contains 37,920 samples and the testing set includes 18,960 samples. In X-Sub, the subjects of the training set and the testing set are different. The training set contains 40,320 videos, and the testing set contains 16,560 videos. 

\textbf{NTU-RGBD 120 (NTU-120):} NTU-120 is an extension of NTU-60 with additional 57,367 skeleton sequences, which totally contains 114,480 samples. The action categories in this dataset can be divided into three major groups: 82 daily actions (eating, sitting down, standing up, etc), 12 health-related actions (falling down, blowing nose, etc), and 26 mutual actions (hugging, shaking hands, etc). Similar to NTU-60, NTU-120 also has two benchmarks: 1) cross-subject (X-Sub), 2) cross-setup (X-Set). In cross-subject, the 106 subjects are equally split into the training set and the testing set. In cross-setup, the samples whose ID is even  belongs to the training set, while the samples whose ID is odd are treated as the testing set. 

\textbf{Northwestern-UCLA (NW-UCLA):} NW-UCLA \cite{81} contains 1,494 video clips covering 10 categories and is captured by three Kinect cameras simultaneously from a variety of viewpoints. Each action sample is performed by 10 different actors. Following \cite{81}, we adopt the same protocol to divide the dataset that the training set is composed of the samples of the first and second viewpoints, while the testing set is made up of samples of the third viewpoint.

\subsection{Implementation Details}
The implementation of our model is conducted on the PyTorch framework. We use the stochastic gradient descent (SGD) as the optimizer to train our network, where the momentum is set to 0.9 and the weight decay is set to 0.001. For NTU-60 and NTU-120, we set the number of training epochs to 75 and the initial learning rate to 0.001. The learning rate decays at the 45th and the 55th epoch. For the NW-UCLA dataset, the number of training epochs is set to 65, and the learning rate decays at the 50th epoch, which is initially set to 0.01. In contrastive learning, the hyper-parameter $\beta$ is set to 0.01 for NTU-60, 0.05 for NTU-120, and 0.1 for NW-UCLA, which is used to control the magnitude of the contrastive learning task. For the cyclical focal loss, we set the cyclical factor $f_c$ to 4 and both hyper-parameters $\gamma_{lc}$ and $\gamma_{hc}$ to 2.

\subsection{Comparison Against the State of the Art}
To verify the effectiveness of our network, we compare our prediction accuracy with the current state-of-the-art methods on NTU-60, NTU-120 and NW-UCLA datasets under two evaluation protocols, including linear evaluation protocol and fine-tuning protocol.

\textbf{Linear Evaluation Protocol.} With this protocol, we evaluate the quality of the representations learned by our method with training a linear classifier (including a fully-connected layer and a softmax layer) and freezing the parameters of the other part. We report the comparison results in Table \ref{1}, Table \ref{2}, and Table \ref{3}. 

\begin{table}[h]  
	\caption{Performance comparison on the NTU-60 dataset with linear evaluation protocol. }
	\centering
	\scalebox{1.2}{
		\renewcommand{\arraystretch}{1.25}
		\begin{tabular}{p{2.5cm}<{\centering}p{1.5cm}<{\centering}p{1.5cm}<{\centering}}  
			\hline
			Methods & X-View(\%) & X-Sub(\%)\\  
			\hline
			
			LongT GAN \cite{75}                             & 39.1  & 48.1 \\ 
			$\rm{MS^2L}$ \cite{77}                               & -     & 52.6 \\ 
			PCRP \cite{78}                                  & 63.5  & 53.9 \\ 
			AS-CAL \cite{79}                                & 64.8  & 58.5 \\ 
			CRRL \cite{70}                                  & 73.8  & 67.6 \\ 
			3s-CrossSCLR \cite{80}                          & 83.4  & 77.8 \\ 
			3s-AimCLR \cite{73}                             & 83.8  & 78.9 \\ 
			BRL \cite{83}                                    & 91.2  & \textbf{86.8} \\ \hline
			\textbf{ConGT}  & \textbf{92.0}  & 86.2 \\ \hline
			\label{1}
	\end{tabular}}
	\vspace{-9mm}  
\end{table}

Table \ref{1} shows the comparisons with previous related methods on NTU-60. Our method achieves the accuracy of 92.0\% on X-View and 86.2\% on X-Sub. Compared with LongT GAN \cite{75}, $\rm{MS^2L}$ \cite{77}, PCRP \cite{78}, and AS-CAL \cite{79}, our method achieves an overwhelming performance. CRRL \cite{70} is a contrast-reconstruction representation learning network that can simultaneously capture postures and motion dynamics for unsupervised  skeleton-based action recognition. By contrast, our method performs 18.2\% better on X-View and 18.6\% on X-Sub. 3s-CrosSCLR \cite{80} also attains superior performance due to its multi-view strategy. Our ConGT achieves excellent performance that outperforms it 8.6\% on X-View and 8.4\% on X-Sub. 3s-AimCLR \cite{73} is a contrastive learning framework with utilizing abundant information mining for self-supervised action representation. Compared with it, the performance of our method is 8.2\% and 7.3\% higher on X-View and X-Sub under Top-1 recognition accuracy, respectively. Compared to other unsupervised methods, BRL \cite{83} gains remarkable results by using the data augmentation and multi-viewpoint sampling strategies, which achieves the accuracy of 91.2\% on X-View and 86.8\% on X-Sub. Our method still works better than it on X-View.

\begin{table}[h]  
	\caption{Performance comparison on the NTU-120 dataset with linear evaluation protocol. }
	\centering
	\scalebox{1.2}{
		\renewcommand{\arraystretch}{1.25}
		\begin{tabular}{p{2.5cm}<{\centering}p{1.5cm}<{\centering}p{1.5cm}<{\centering}}  
			\hline
			Methods & X-Set(\%) & X-Sub(\%)\\  
			\hline
			
			LongT GAN \cite{75}                             & 39.7  & 35.6 \\ 
			PCRP \cite{78}                                  & 45.1  & 41.7 \\ 
			AS-CAL \cite{79}                                & 49.2  & 48.6 \\
			CRRL \cite{70}                                  & 57.0  & 56.2 \\ 
			3s-CrossSCLR \cite{80}                          & 66.7  & 67.9 \\ 
			3s-AimCLR \cite{73}                             & 68.8  & 68.2 \\
			ISC \cite{84}                                   & 67.1  & 67.9 \\ 
			BRL \cite{83}                                   & 79.2  & 77.1 \\ \hline
			\textbf{ConGT}  & \textbf{80.5}  & \textbf{78.6} \\ \hline
			\label{2}
	\end{tabular}}
	\vspace{-9mm}  
\end{table}

In Table \ref{2}, we conduct the comparative experiment on NTU-120 dataset on both X-Sub and X-Set benchmarks. We also follow the standard practice in the literature, reporting the top-1 classification accuracies on both benchmarks. The competitive results in Table \ref{2} verify the superiority of our proposed method over all methods. 

\begin{figure}[htbp]
	\centering  
	\includegraphics [scale=0.6]{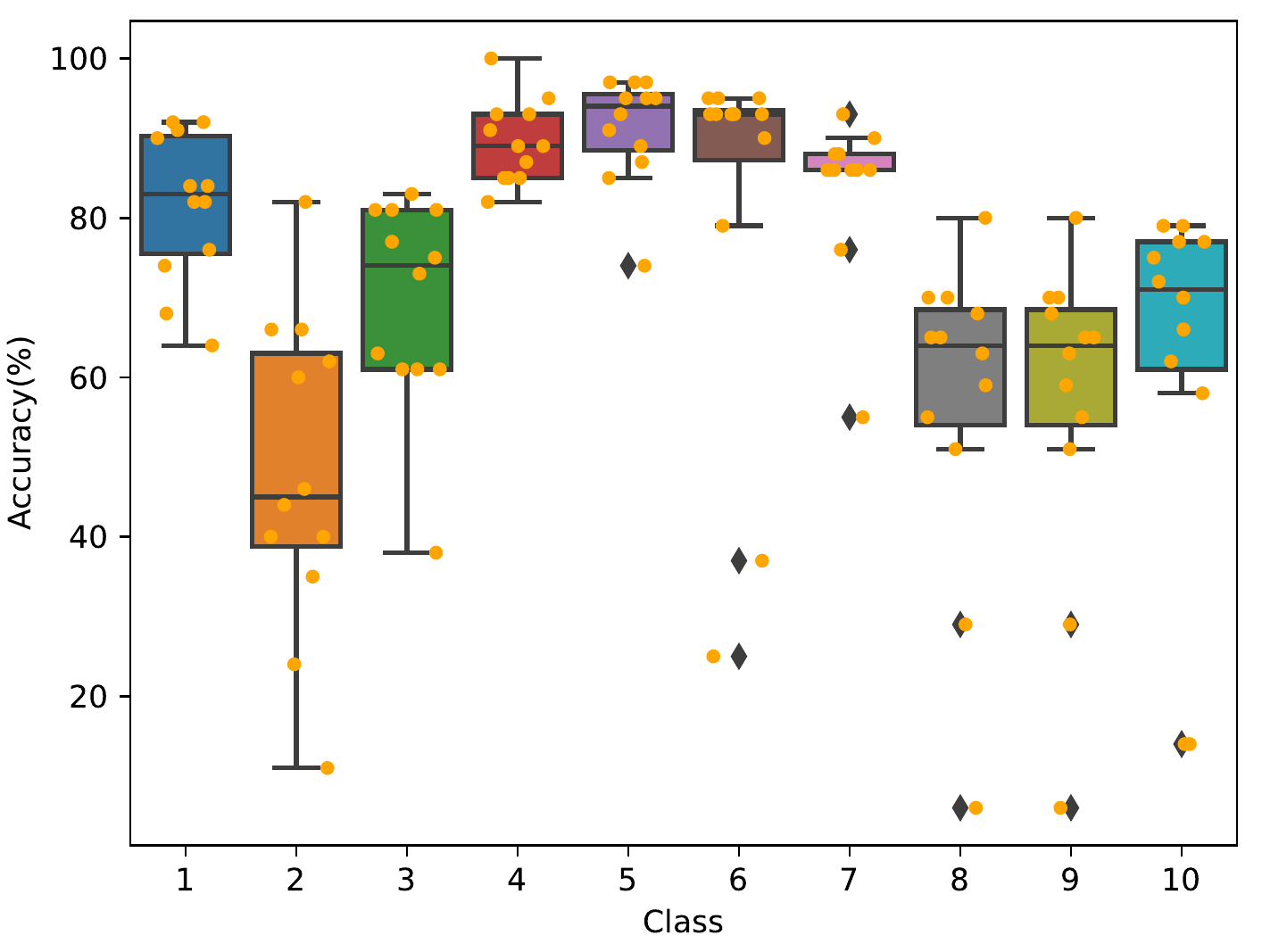}
	
	\caption  {Different color boxes indicate the accuracy range of several categories, the black line inside each box represents the median value, boxes limits include interquartile ranges from 25\% to 75\% of samples, upper and lower whiskers are computed as 1.5 times the distance of upper and lower limits of the box, and all values outside the whiskers are considered as outliers.}
	\label{box}
\end{figure}

\begin{table}[h]  
	\caption{Performance comparison on the NW-UCLA dataset with linear evaluation protocol.}
	\centering
	\scalebox{1.2}{
		\renewcommand{\arraystretch}{1.25}
		\begin{tabular}{p{3cm}<{\centering}p{1.5cm}<{\centering}p{1.5cm}<{\centering}}  
			\hline
			Methods & Accuracy(\%) \\  
			\hline
			
			LongT GAN \cite{75}                             & 74.3  \\ 
			$\rm{MS^2L}$ \cite{77}                                  & 76.8 \\ 
			CRRL \cite{70}                                  & 83.8   \\ \hline
			\textbf{ConGT}  & \textbf{85.3} \\ \hline
			\label{3}
	\end{tabular}}
	\vspace{-9mm}  
\end{table}

As shown in Table \ref{3}, the proposed ConGT achieves the best accuracy of 85.3\% on the NW-UCLA dataset, surpassing the previous state-of-the-art methods. The NW-UCLA contains ten categories of actions: pick up with on hand, pick up with two hands, drop trash, walk around, sit down, stand up, donning, doffing, throw, and carry. For each specific action, we use the boxplots to show the training accuracy of every 5 epochs in Fig. \ref{box}, where these ten classes are denoted by numbers 1-10.

\begin{table}[h]  
	\caption{Performance comparison on the NTU-60 and NTU-120 dataset with fine-tuning protocol.}
	\centering
	\scalebox{1.15}{
		\renewcommand{\arraystretch}{1.25}
		\begin{tabular}{c|c|c|c|c}  
			\hline
			\multirow{2}{*}{Methods} & \multicolumn{2}{c|}{NTU-60} & \multicolumn{2}{c}{NTU-120}  \cr
			\cline{2-5}
			&X-View &X-Sub & X-Set & X-Sub
			\\
			\hline
			
			SkeletonCLR \cite{80}                          & 88.9  & 82.2 &75.3 &73.6 \\ 
			AimCLR \cite{73}                                & 89.2  & 83.0 &76.7 &76.4\\ \hline
			\textbf{ConGT}  & \textbf{91.6}  & \textbf{84.6} & \textbf{80.5} &\textbf{79.4}\\ \hline
			
	\end{tabular}}
	\label{4}
\end{table}

\textbf{Fine-tuning Protocol.} Following \cite{80}, we first pre-train STG, STT and contrastive learning and then append a linear classifier to retrain the whole model, where the parameters of each layer in our network are updated with the backpropagation. We compare our model with the state-of-the-art methods in Table \ref{4}. 

To make a fair comparison with SkeletonCLR \cite{80} and AimCLR \cite{73}, we only use the bone data to compare the finetuned results. As shown in Table \ref{4}, our ConGT defeats them both on NTU-60 and NTU-120. Specifically, on X-View and X-Sub of NTU-60, our method surpasses SkeletonCLR by 2.7\% and 2.4\%, respectively, and outperforms AimCLR by 2.4\% and 1.6\%, respectively. On NTU-120, compared to SkeletonCLR, the improvements reach 5.2\% and 5.8\% on X-Set and X-Sub, respectively. Compared to AimCLR, our method surpasses it by 3.8\% and 3.0\% on X-Set and X-Sub, respectively. The results demonstrate that our model with the contrastive learning paradigm can effectively learn rich action representations of human actions.

\subsection{Ablation Study}
In this section, we design ablation experiments to investigate the effectiveness of the proposed approach. We first validate the effectiveness of each component of our model. And we demonstrate that the existence of over-smoothing problem during the accumulation of GCN layers. Then we test the influence of the hyper-parameter $\beta$ that controls the magnitude of contrastive learning. Finally, we verify the effectiveness of the cyclical focal loss. 

\begin{table}[h]  
	\caption{The comparison performance of ConGT with different parts on X-View of NTU-60.}
	\centering
	\scalebox{1.2}{
		\renewcommand{\arraystretch}{1.25}
		\begin{tabular}{c|c|c|c|c}  
			\hline
			Subnet & STG & STT &CL &Accuracy(\%) \\ \hline
			
			N-STG      & \checkmark  &            &        &89.6 \\ \hline 
			N-STT      &             & \checkmark &        &32.5 \\ \hline
			ST-GT    & \checkmark  & \checkmark &        &73.3 \\ \hline
			ST-MGT   & \checkmark  & \checkmark &        &90.2 \\ \hline
			\textbf{ConGT} & \checkmark  & \checkmark & \checkmark  & \textbf{92.4} \\ \hline

	\end{tabular}}
	\label{5}
\end{table}

\subsubsection{Impact of Each Component in ConGT}
As three primary components of our proposed ConGT, the Spatial-Temporal Graph Convolution stream (STG), Spatial-Temporal Transformer stream (STT), and Contrastive Learning (CL) are also the main contributions in this work. To evaluate the effectiveness of STG, STT and CL, we design four subnets. 
\begin{itemize}
	\item[$\bullet$] \textbf{N-STG}: Only training STG to show the results of using GCN alone.
	\item[$\bullet$] \textbf{N-STT}: Only training STT to display the results of using transformer alone.
	\item[$\bullet$] \textbf{ST-GT}: We remove the contrastive learning part and add the representations learned by the STG and STT to demonstrate the effectiveness of contrastive learning.
	\item[$\bullet$] \textbf{ST-MGT}: We  replace the InfoNCE loss with the MSE loss to illustrate that the contrastive learning plays a crucial role in combining two different types of action representations.
\end{itemize}

On the four subnets, we conduct experiments on X-View of NTU-60. The results obtained by these baselines are depicted in Table \ref{5}. It can intuitively see that the recognition accuracy can reach 89.6\% when only using the STG, while the recognition accuracy is only 32.5\% when only using the STT. We speculate the reason is that the transformer treats each node as a separate unit and regards the human skeleton as a complete graph with connections built between each joint and the rest joints, resulting in less variation between different movements. 

\begin{figure}
	\centering  
	\includegraphics [scale=0.32]{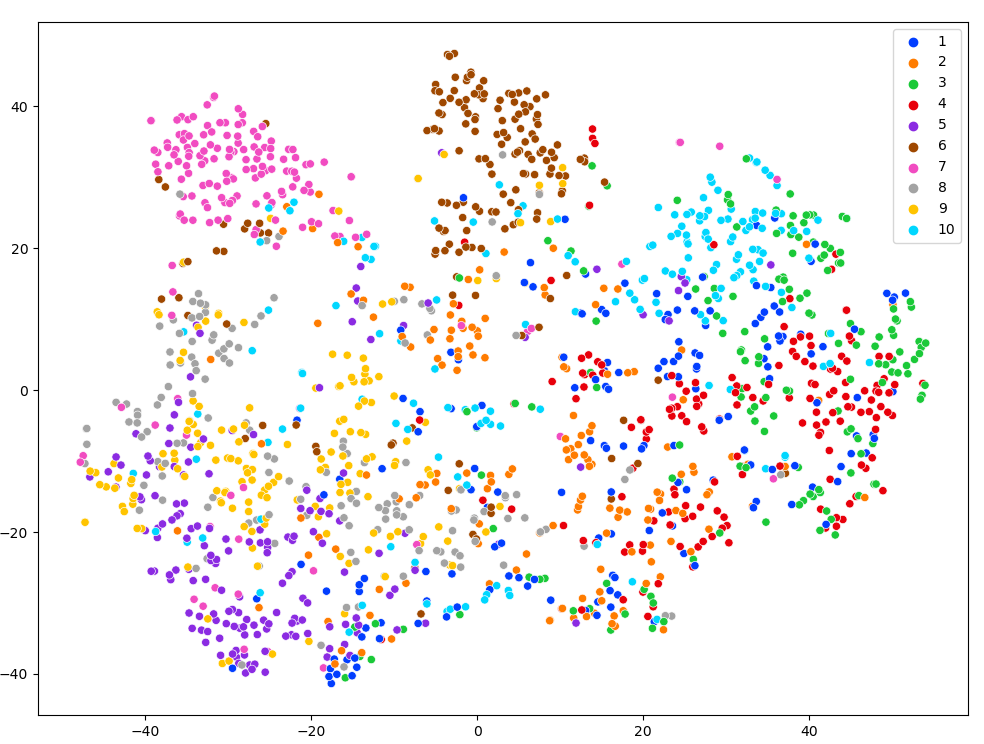}
	
	\caption  {The visualization of the action representation learned by STT. In order to better show the distribution of action representation of each category, we select the top 10 classes of the X-View of NTU-60 dataset. Each color denotes an action class and each point represents a skeleton sequence. }
	\label{transformer}
\end{figure}

\begin{figure}[h]
	\centering  
	\includegraphics [scale=0.32]{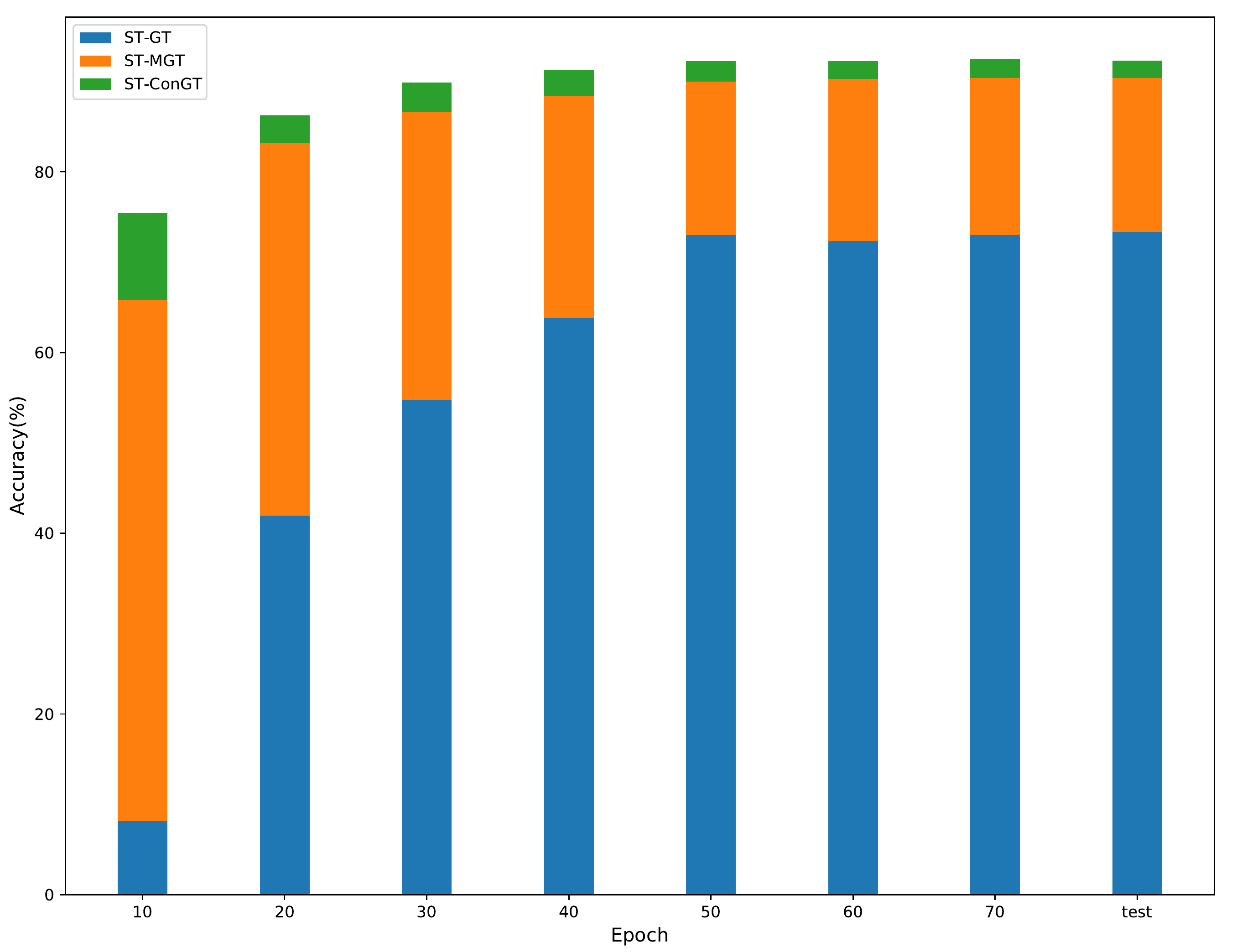}
	
	\caption  {The influence of the contrastive learning paradigm. The blue part denotes the recognition accuracy of ST-GT. The orange part represents the improvement of the recognition accuracy with the MSE loss. The green part shows the superiority of using contrastive learning. }
	\label{histogram}
\end{figure}

\begin{figure*}[h]
	\centering
	\begin{subfigure}{2.35in}
		\centering
		\includegraphics[scale=0.22]{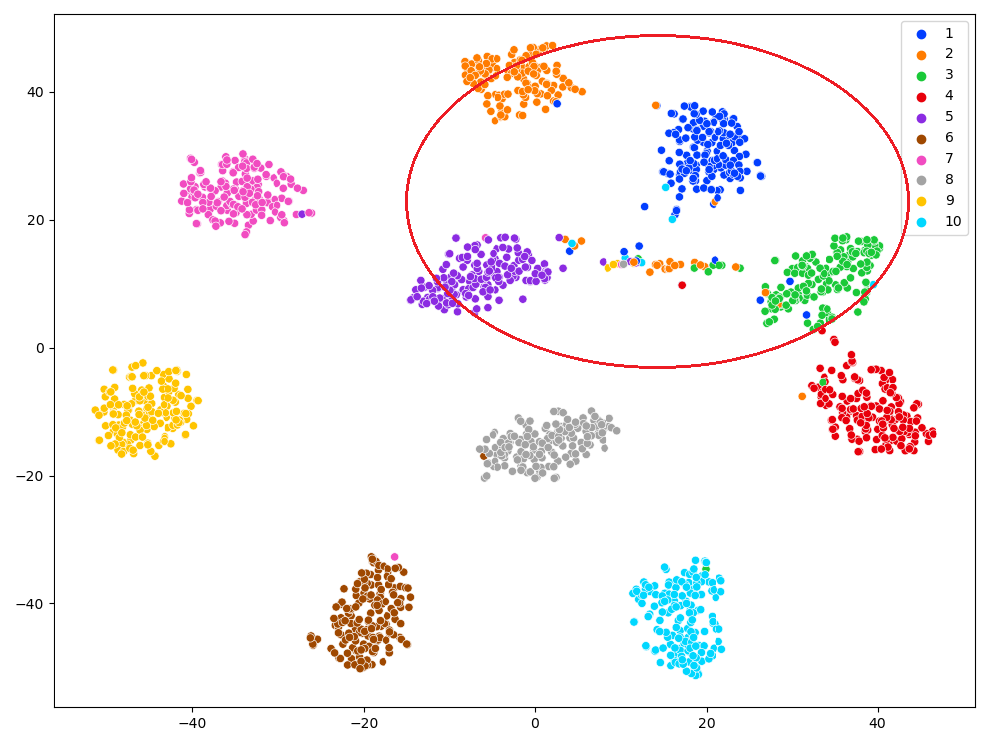}
		\caption{}  	
	\end{subfigure}
	\begin{subfigure}{2.35in}
		\centering
		\includegraphics[scale=0.22]{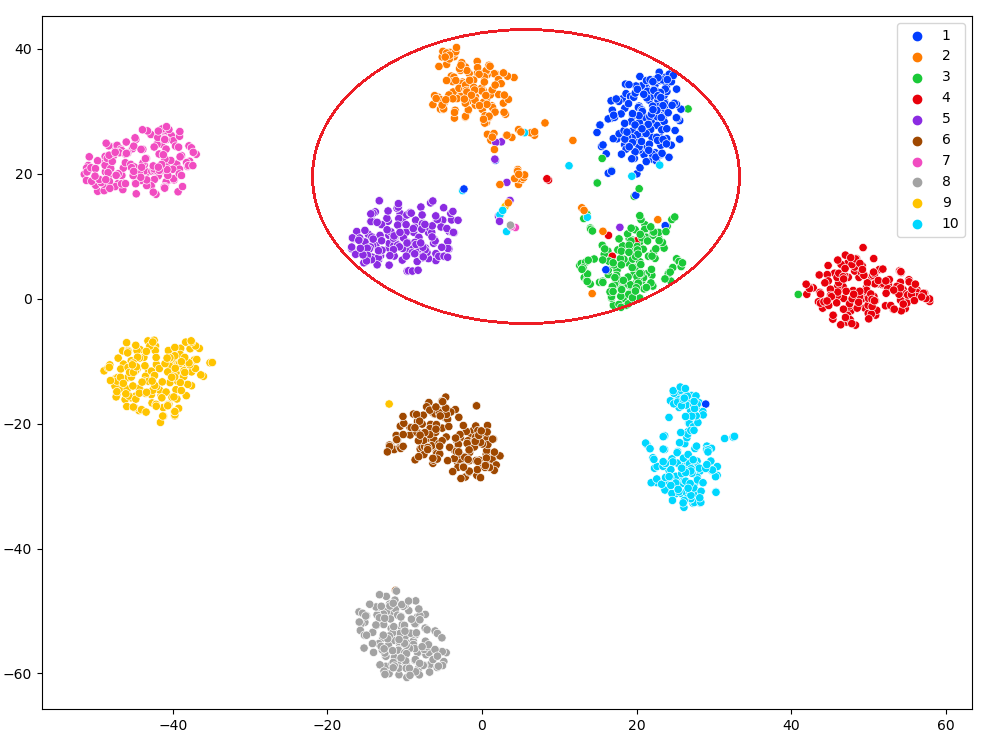}
		\caption{}
	\end{subfigure}
	\begin{subfigure}{2.35in}
		\centering
		\includegraphics[scale=0.22]{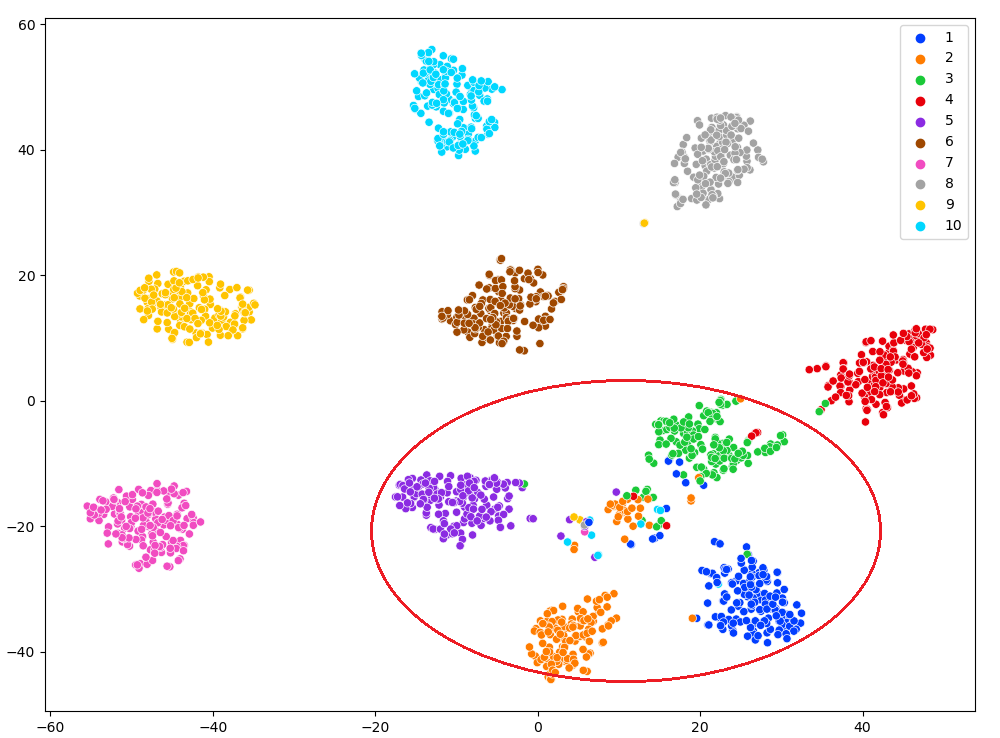}
		\caption{}
	\end{subfigure}
	\caption{The t-SNE visualization of action representations. Each point represents a skeleton sequence. We show the first 10 action classes of the X-View of NTU-60 dataset, indicated by colors. (a). The STG with 6 GCN layers. (b). The STG with 9 GCN layers. (c) The STG with 12 GCN layers.}\label{gcn-10}
\end{figure*}

\begin{figure*}
	\centering
	\begin{subfigure}{2.35in}
		\centering
		\includegraphics[scale=0.4]{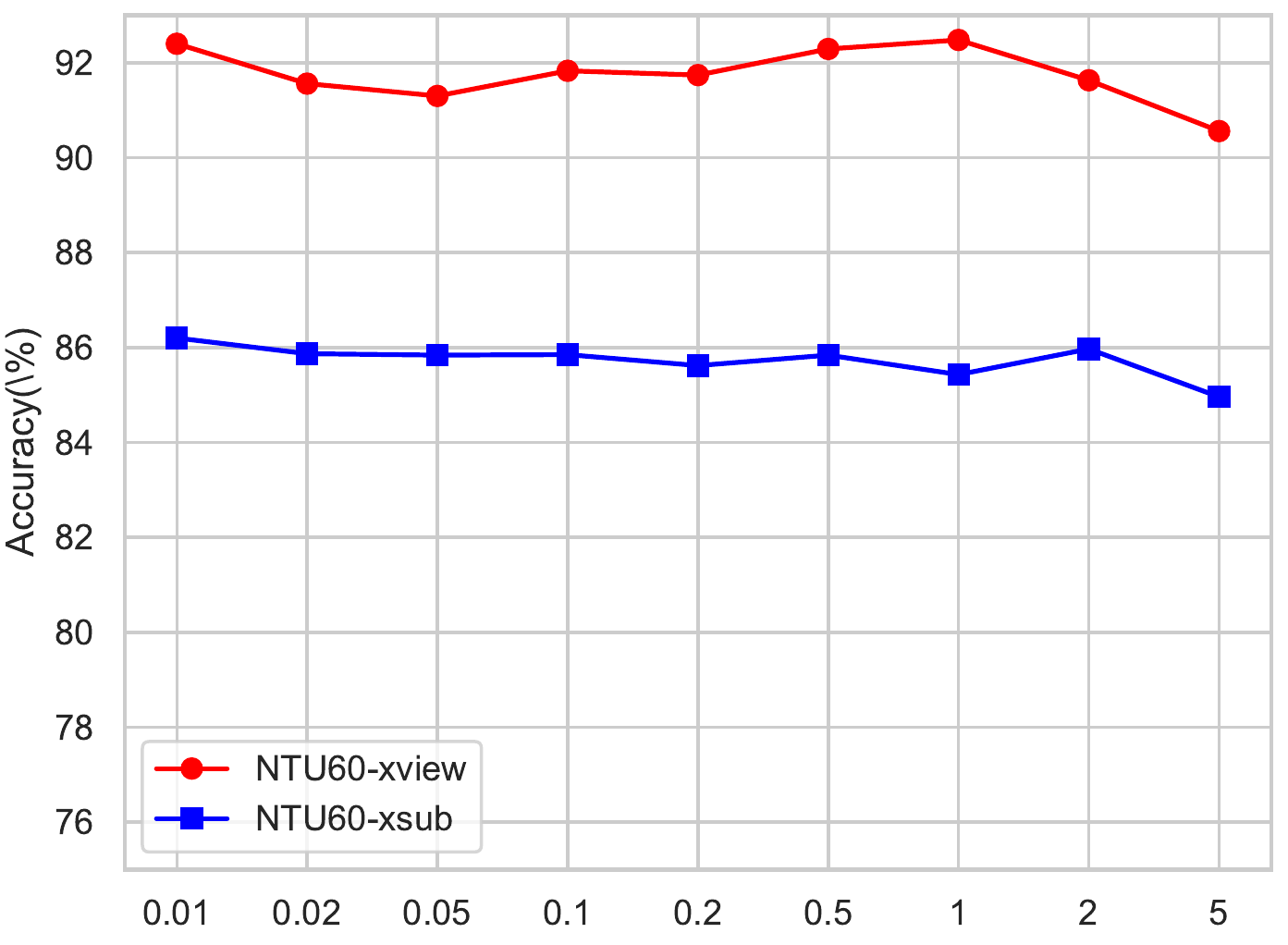}
		\caption{}  	
	\end{subfigure}
	\begin{subfigure}{2.35in}
		\centering
		\includegraphics[scale=0.42]{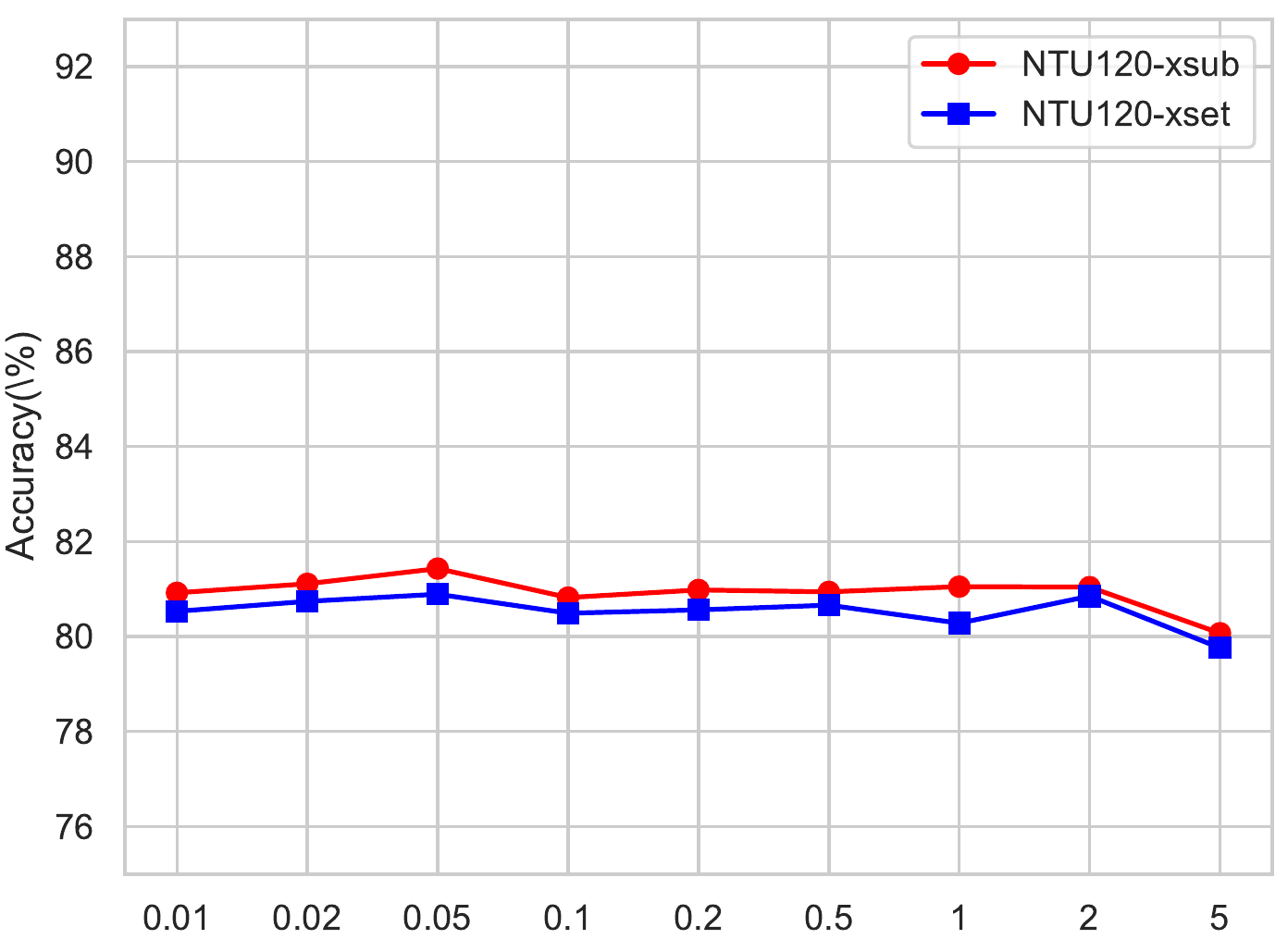}
		\caption{}
	\end{subfigure}
	\begin{subfigure}{2.35in}
		\centering
		\includegraphics[scale=0.41]{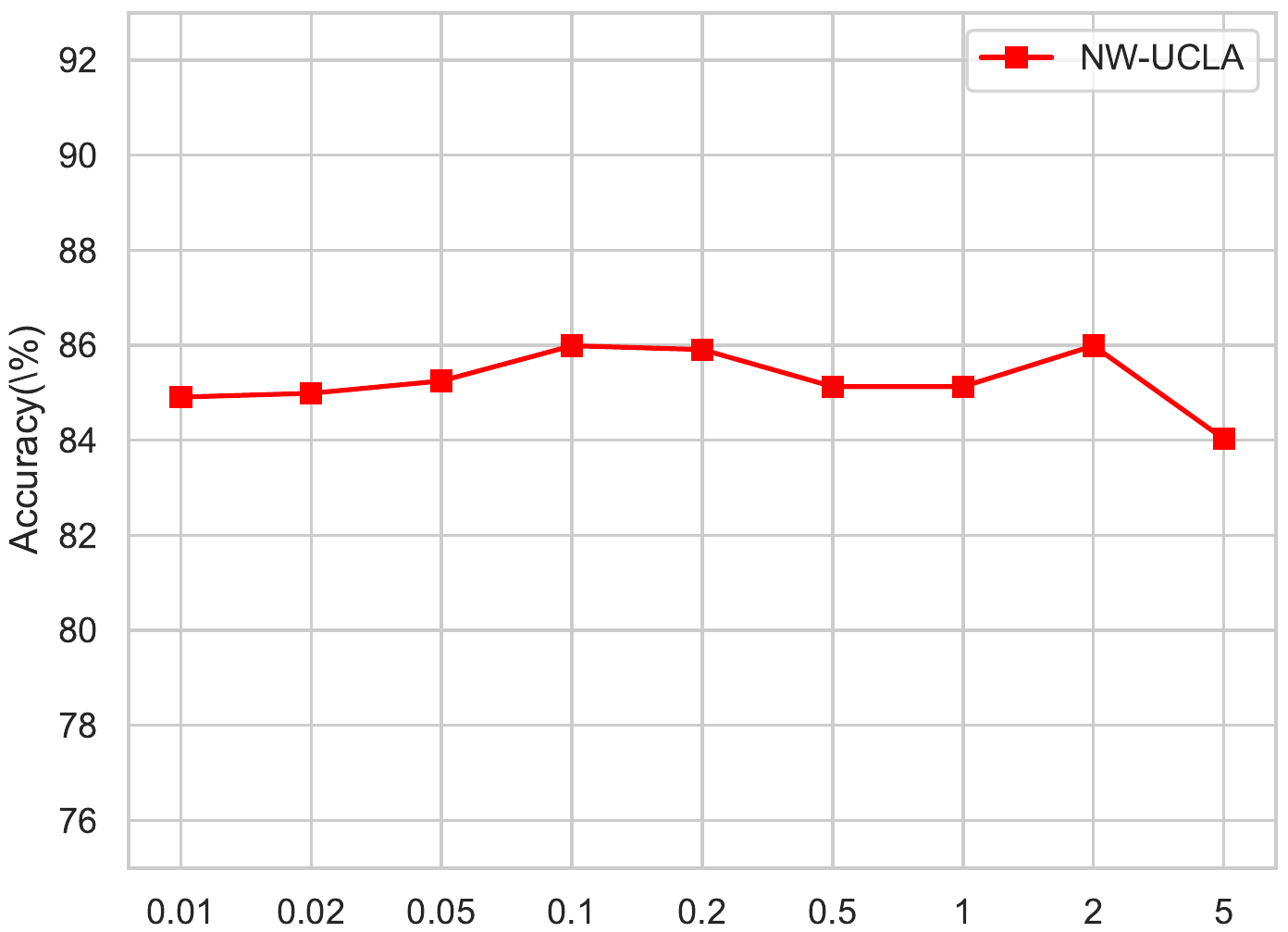}
		\caption{}
	\end{subfigure}
	\caption{The influence of the magnitude of contrastive learning on (a) NTU-60, (b) NTU-120, and (c) NW-UCLA.}\label{line}
\end{figure*}

To verify this claim, we visualize the action representation learned by STT in Fig. \ref{transformer}, where all the categories are mixed together. This adds to the evidence that it is unreasonable to treat the human skeleton as a complete graph.

Moreover, we adopt two different methods to demonstrate the effectiveness of contrastive learning in Table \ref{5}. In ST-GT, we add the action representations output by STG and STT together. In this way, the accuracy is reduced by 19.1\% compared to ConGT. Furthermore, we replace the InfoNCE loss with MSE loss to evaluate the effect of contrastive learning in ST-MGT and the final recognition accuracy has a decline. In Fig. \ref{histogram}, we depict the results of the ST-GT, ST-MGT, and ConGT. The blue part denotes the recognition accuracy of ST-GT, the orange part represents the improvement of the ST-MGT using the MSE loss, and the green part shows the superiority of using contrastive learning in ConGT. We can intuitively see that the training accuracy of ST-GT and ST-MGT are consistently lower than ConGT. Therefore, it can be illustrated that the contrastive learning plays a crucial role in fusing long-distance relationships into the  topology structure of the human skeleton graph.

\begin{table}[h]  
	\caption{Comparison of the performance (accuracy (\%)) on the X-View setting of the NTU-60 dataset when the STG with or without the adjacency matrix $A$, the learnable matrix $L$, and the embedding matrix $E$. wo/$A$ denotes without $A$, wo/$L$ denotes without $L$, and wo/$E$ denotes without $E$.
	}
	\centering
	\scalebox{1.2}{
		\renewcommand{\arraystretch}{1.25}
		\begin{tabular}{p{2cm}<{\centering}|p{3cm}<{\centering}}  
			\hline
			Methods  &Accuracy(\%) \\ \hline
			
			ST-GCN      & 88.3 \\ \hline 
			STG         & 89.2 \\ \hline
			STG wo/$A$    & 88.5 \\ \hline
			STG wo/$L$    & 88.3 \\ \hline
			STG wo/$E$    & 88.7 \\ \hline
			
	\end{tabular}}
	\label{6}
\end{table}

\subsubsection{The Effect of Adaptive Graph Strategy in STG}
We demonstrate the influence of AGCN by comparing STG with ST-GCN in the supervised manner. For a fair comparison, we set the same number of GCN layers in STG as that in ST-GCN. In Table \ref{6}, we can see that the recognition accuracy of the STG outperforms 0.9\% that of ST-GCN, which indicates that the adaptive graph strategy contributes to improving the accuracy of action recognition. Furthermore, to evaluate the necessity of the three graphs in AGCN, we conduct an ablation study on the three graphs. We manually delete one of the three types of graphs and show their performance in Table \ref{6}. We find that taking away any one of the three graphs will affect the final recognition result negatively. When all three graphs are simultaneously enabled, our model can achieve the best performance. This indicates that the adaptive graph strategy is conducive to increasing the accuracy of action recognition.

\begin{table}[h]  
	\caption{Recognition accuracies obtained by STG containing 6, 9, and 12 GCN layers on X-View setting of NTU-60.	}
	\centering
	\scalebox{1.2}{
		\renewcommand{\arraystretch}{1.25}
		\begin{tabular}{p{1cm}<{\centering}|p{3cm}<{\centering}}  
			\hline
			Layers  &Accuracy(\%) \\ \hline
			
			6      & 89.6 \\ \hline 
			9      & 89.2 \\ \hline
			12     & 86.1 \\ \hline
			
	\end{tabular}}
	\label{7}
\end{table}

\subsubsection{The Impact of the Number of GCN Layers in STG}
In addition, to demonstrate that the over-smoothing problem occurs during the accumulation of GCN layers, we compare the recognition performance of STG containing 6, 9, and 12 GCN layers on the X-View setting of NTU-60. In Table \ref{7}, we can see that the recognition accuracy has a significant decline when the GCN layer number is increased to 12. We also apply t-SNE \cite{86} to show the embedding distribution of these three options in Fig. \ref{gcn-10}. From the visual results, we can find that with the increase of network layers, the action representations of classes 1, 2, 3, and 5 (circled by the red ellipse) tend to be consistent. It also reveals that with the increase of the number of GCN layers, the probability of the over-smoothing problem also increases.

\begin{table}[h]  
	\caption{Comparison of the top-1 test recognition accuracies for Cross-Entropy Loss and Cyclical Focal Loss on NTU-60 and NTU-120.}
	\centering
	\scalebox{1}{
		\renewcommand{\arraystretch}{1.25}
		\begin{tabular}{p{2.5cm}<{\centering}|p{2.5cm}<{\centering}|p{2.5cm}<{\centering}}  
			\hline
			Datasets  & Cross-Entropy Loss & Cyclical Focal Loss \\ \hline
			
			NTU-60 (X-View)      & 91.08  & 91.59 \\ \hline 
			NTU-60 (X-Sub)       & 84.21  & 84.55 \\ \hline
			NTU-120 (X-Set)      & 78.80  & 80.53 \\ \hline
			NTU-120 (X-Sub)      & 78.82  & 79.36\\ \hline
			
	\end{tabular}}
	\label{8}
\end{table}

\subsubsection{The Impact of the Contrastive Learning Hyper-parameter}
To justify the impact of the hyper-parameter $\beta$ on controlling the magnitude of contrastive learning, we examine the performance of ConGT with a set of representative $\beta$ values \{0.01, 0.02, 0.05, 0.1, 0.2, 0.5, 1, 2, 5\}. 

The performance results are shown in Fig. \ref{line}. As we can see, our model performs best on NTU-60 when $\beta$ is 0.01. While for NTU-120, the best $\beta$ is 0.05. When $\beta$ is 0.1, our method achieves the best accuracy on NW-UCLA. In addition, it can be seen that when $\beta$ becomes large, the performance of ConGT on both NTU-60, NTU-120, and NW-UCLA will decline. We suspect it is due to that the gradient conflict between the action recognition task and the contrastive task.  Therefore, it is necessary to select an appropriate $\beta$, when involving the contrastive learning paradigm.

\subsubsection{The Effectiveness of the Cyclical Focal Loss}
In this section, we compare the recognition results obtained by using the cross-entropy loss and the cyclical focal loss on NTU-60 and NTU-120 in Table \ref{8}. It shows that when the model is trained with the cyclical focal loss, the test accuracy is consistently better than that using the cross-entropy loss.

\begin{figure}[h]
	\centering  
	\includegraphics [scale=0.32]{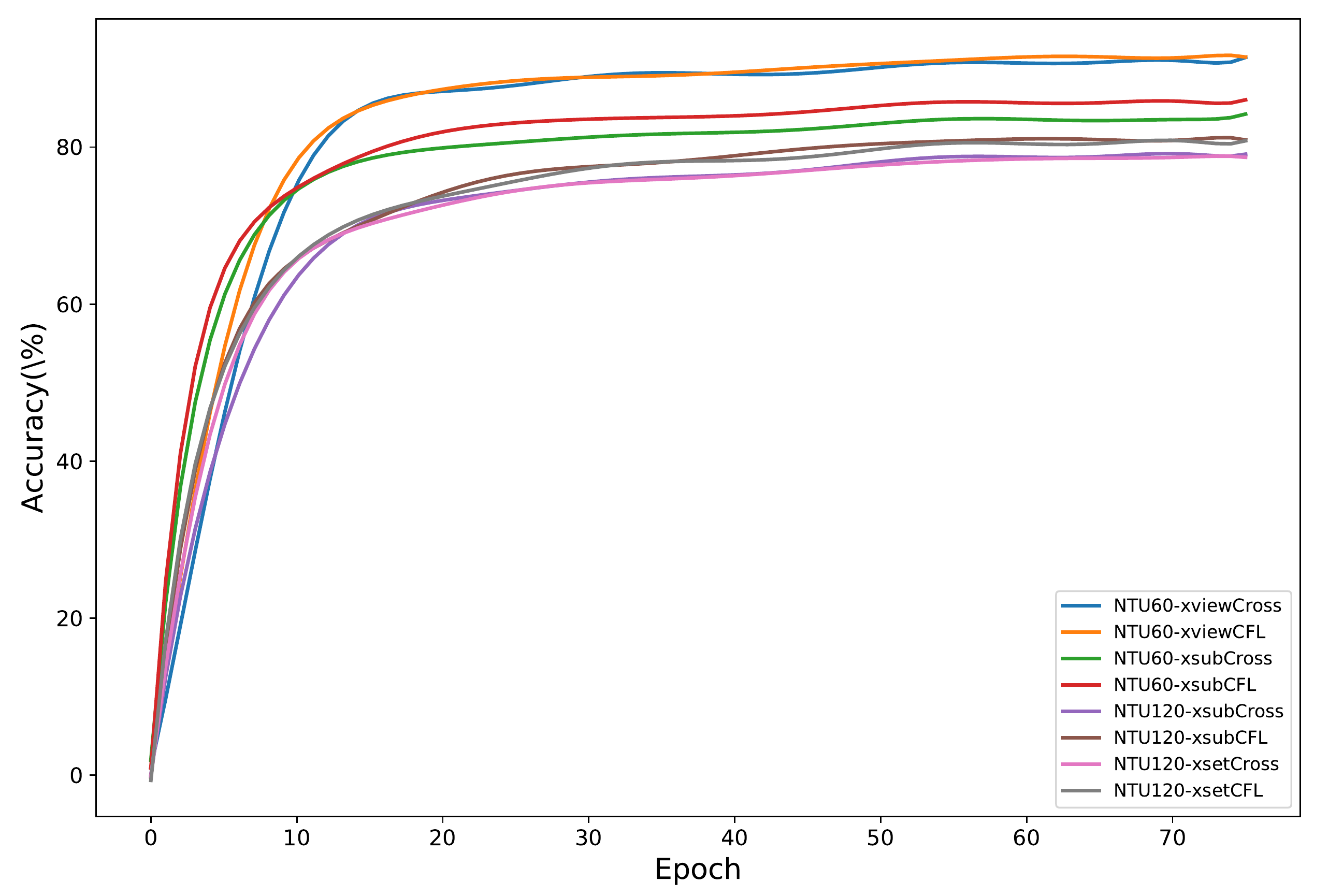}
	
	\caption  {The accuracy curves of training ConGT using cross-entropy loss and cyclical focal loss. }
	\label{loss}
\end{figure}

Fig. \ref{loss} shows the training accuracy curves of training our network using the cross-entropy loss and the cyclical focal loss. Although the cross-entropy loss curve and the cyclical focal loss curve on the same dataset have strong similarities, it is notable that training with the cyclical focal loss provides a slightly faster learning convergence in the training. Therefore, it can be confirmed that the cyclical focal loss better helps the learning in the early epochs.

\section{Conclusion}
In this work, we design a novel Contrastive GCN-Transformer Network (ConGT), which can capture the relationships between arbitrary joints in the intra- and inter- frames more accurately while maintaining the topology structure of human skeleton graphs. Specifically, the STG is designed to obtain action representations maintaining the topology structure of the human skeleton graph. At the same time, the STT is used to acquire action representations containing the global relationships among joints. Moreover, we introduce the contrastive learning paradigm, serving as an auxiliary task, to maximize the mutual information between the action representations learned via the two streams to improve the action recognition task. In this manner, we can make up for the weak ability of GCN to capture long-distance features on the basis of maintaining the topology structure of the human skeleton graph and reduce the risk of network over-smoothing. In addition, we introduce the cyclical focal loss as the learning objective of our model, which places heavy weights on confident training samples in the first training epochs of a neural network. Ablation studies have been performed in this work, which verify the effectiveness of our method. Experiments on three publicly available datasets demonstrate the superiority of our proposed method over other methods.  

\section*{Acknowledgment}
This work was supported in part by the National Natural Science Foundation of China (No. 61976127), Shandong Provincial Natural Science Foundation (Nos. ZR2021LZL012, ZR2021QG004).

\bibliographystyle{IEEEtran}
\bibliography{IEEEexample}

\begin{thebibliography}{10}
\providecommand{\url}[1]{#1}
\csname url@samestyle\endcsname
\providecommand{\newblock}{\relax}
\providecommand{\bibinfo}[2]{#2}
\providecommand{\BIBentrySTDinterwordspacing}{\spaceskip=0pt\relax}
\providecommand{\BIBentryALTinterwordstretchfactor}{4}
\providecommand{\BIBentryALTinterwordspacing}{\spaceskip=\fontdimen2\font plus
\BIBentryALTinterwordstretchfactor\fontdimen3\font minus
  \fontdimen4\font\relax}
\providecommand{\BIBforeignlanguage}[2]{{%
\expandafter\ifx\csname l@#1\endcsname\relax
\typeout{** WARNING: IEEEtran.bst: No hyphenation pattern has been}%
\typeout{** loaded for the language `#1'. Using the pattern for}%
\typeout{** the default language instead.}%
\else
\language=\csname l@#1\endcsname
\fi
#2}}
\providecommand{\BIBdecl}{\relax}
\BIBdecl

\bibitem{6}
Y.~Ming, F.~Feng, C.~Li, and J.-H. Xue, ``3d-tdc: A 3d temporal dilation
  convolution framework for video action recognition,'' \emph{Neurocomputing},
  vol. 450, pp. 362--371, 2021.

\bibitem{7}
I.~Rodr{\'\i}guez-Moreno, J.~M. Mart{\'\i}nez-Otzeta, I.~Goienetxea,
  I.~Rodriguez-Rodriguez, and B.~Sierra, ``Shedding light on people action
  recognition in social robotics by means of common spatial patterns,''
  \emph{Sensors}, vol.~20, no.~8, p. 2436, 2020.

\bibitem{8}
Z.~Xu, G.~Wang, and X.~Guo, ``Sensor-based activity recognition of solitary
  elderly via stigmergy and two-layer framework,'' \emph{Engineering
  Applications of Artificial Intelligence}, vol.~95, p. 103859, 2020.

\bibitem{24}
Q.~Ke, M.~Bennamoun, S.~An, F.~Sohel, and F.~Boussaid, ``A new representation
  of skeleton sequences for 3d action recognition,'' in \emph{Proceedings of
  the IEEE conference on computer vision and pattern recognition}, 2017, pp.
  3288--3297.

\bibitem{25}
P.~Wang, Z.~Li, Y.~Hou, and W.~Li, ``Action recognition based on joint
  trajectory maps using convolutional neural networks,'' in \emph{Proceedings
  of the 24th ACM international conference on Multimedia}, 2016, pp. 102--106.

\bibitem{29}
Y.~Du, W.~Wang, and L.~Wang, ``Hierarchical recurrent neural network for
  skeleton based action recognition,'' in \emph{Proceedings of the IEEE
  conference on computer vision and pattern recognition}, 2015, pp. 1110--1118.

\bibitem{31}
J.~Liu, A.~Shahroudy, D.~Xu, and G.~Wang, ``Spatio-temporal lstm with trust
  gates for 3d human action recognition,'' in \emph{European conference on
  computer vision}.\hskip 1em plus 0.5em minus 0.4em\relax Springer, 2016, pp.
  816--833.

\bibitem{13}
S.~Yan, Y.~Xiong, and D.~Lin, ``Spatial temporal graph convolutional networks
  for skeleton-based action recognition,'' in \emph{Thirty-second AAAI
  conference on artificial intelligence}, 2018.

\bibitem{14}
L.~Shi, Y.~Zhang, J.~Cheng, and H.~Lu, ``Two-stream adaptive graph
  convolutional networks for skeleton-based action recognition,'' in
  \emph{Proceedings of the IEEE/CVF conference on computer vision and pattern
  recognition}, 2019, pp. 12\,026--12\,035.

\bibitem{39}
M.~Li, S.~Chen, X.~Chen, Y.~Zhang, Y.~Wang, and Q.~Tian, ``Actional-structural
  graph convolutional networks for skeleton-based action recognition,'' in
  \emph{Proceedings of the IEEE/CVF conference on computer vision and pattern
  recognition}, 2019, pp. 3595--3603.

\bibitem{40}
B.~Li, X.~Li, Z.~Zhang, and F.~Wu, ``Spatio-temporal graph routing for
  skeleton-based action recognition,'' in \emph{Proceedings of the AAAI
  Conference on Artificial Intelligence}, vol.~33, no.~01, 2019, pp.
  8561--8568.

\bibitem{48}
{Shi, Lei and Zhang, Yifan and Cheng, Jian and Lu, Hanqing}, ``Decoupled
  spatial-temporal attention network for skeleton-based action recognition,''
  \emph{arXiv preprint arXiv:2007.03263}, 2020.

\bibitem{17}
C.~Plizzari, M.~Cannici, and M.~Matteucci, ``Skeleton-based action recognition
  via spatial and temporal transformer networks,'' \emph{Computer Vision and
  Image Understanding}, vol. 208, p. 103219, 2021.

\bibitem{23}
C.~Li, Q.~Zhong, D.~Xie, and S.~Pu, ``Co-occurrence feature learning from
  skeleton data for action recognition and detection with hierarchical
  aggregation,'' \emph{arXiv preprint arXiv:1804.06055}, 2018.

\bibitem{26}
Y.~Li, R.~Xia, X.~Liu, and Q.~Huang, ``Learning shape-motion representations
  from geometric algebra spatio-temporal model for skeleton-based action
  recognition,'' in \emph{2019 IEEE International Conference on Multimedia and
  Expo (ICME)}.\hskip 1em plus 0.5em minus 0.4em\relax IEEE, 2019, pp.
  1066--1071.

\bibitem{27}
H.~Duan, Y.~Zhao, K.~Chen, D.~Lin, and B.~Dai, ``Revisiting skeleton-based
  action recognition,'' in \emph{Proceedings of the IEEE/CVF Conference on
  Computer Vision and Pattern Recognition}, 2022, pp. 2969--2978.

\bibitem{28}
W.~Zaremba, I.~Sutskever, and O.~Vinyals, ``Recurrent neural network
  regularization,'' \emph{arXiv preprint arXiv:1409.2329}, 2014.

\bibitem{30}
I.~Lee, D.~Kim, S.~Kang, and S.~Lee, ``Ensemble deep learning for
  skeleton-based action recognition using temporal sliding lstm networks,'' in
  \emph{Proceedings of the IEEE international conference on computer vision},
  2017, pp. 1012--1020.

\bibitem{312}
J.~Liu, G.~Wang, P.~Hu, L.-Y. Duan, and A.~C. Kot, ``Global context-aware
  attention lstm networks for 3d action recognition,'' in \emph{Proceedings of
  the IEEE conference on computer vision and pattern recognition}, 2017, pp.
  1647--1656.

\bibitem{32}
M.~Niepert, M.~Ahmed, and K.~Kutzkov, ``Learning convolutional neural networks
  for graphs,'' in \emph{International conference on machine learning}.\hskip
  1em plus 0.5em minus 0.4em\relax PMLR, 2016, pp. 2014--2023.

\bibitem{33}
J.~Zhu, W.~Zou, Z.~Zhu, and Y.~Hu, ``Convolutional relation network for
  skeleton-based action recognition,'' \emph{Neurocomputing}, vol. 370, pp.
  109--117, 2019.

\bibitem{34}
F.~Monti, D.~Boscaini, J.~Masci, E.~Rodola, J.~Svoboda, and M.~M. Bronstein,
  ``Geometric deep learning on graphs and manifolds using mixture model cnns,''
  in \emph{Proceedings of the IEEE conference on computer vision and pattern
  recognition}, 2017, pp. 5115--5124.

\bibitem{35}
M.~Defferrard, X.~Bresson, and P.~Vandergheynst, ``Convolutional neural
  networks on graphs with fast localized spectral filtering,'' \emph{Advances
  in neural information processing systems}, vol.~29, 2016.

\bibitem{36}
C.~Wang, B.~Samari, and K.~Siddiqi, ``Local spectral graph convolution for
  point set feature learning,'' in \emph{Proceedings of the European conference
  on computer vision (ECCV)}, 2018, pp. 52--66.

\bibitem{41}
L.~Shi, Y.~Zhang, J.~Cheng, and H.~Lu, ``Skeleton-based action recognition with
  directed graph neural networks,'' in \emph{Proceedings of the IEEE/CVF
  Conference on Computer Vision and Pattern Recognition}, 2019, pp. 7912--7921.

\bibitem{42}
Z.~Liu, H.~Zhang, Z.~Chen, Z.~Wang, and W.~Ouyang, ``Disentangling and unifying
  graph convolutions for skeleton-based action recognition,'' in
  \emph{Proceedings of the IEEE/CVF conference on computer vision and pattern
  recognition}, 2020, pp. 143--152.

\bibitem{43}
A.~Vaswani, N.~Shazeer, N.~Parmar, J.~Uszkoreit, L.~Jones, A.~N. Gomez,
  {\L}.~Kaiser, and I.~Polosukhin, ``Attention is all you need,''
  \emph{Advances in neural information processing systems}, vol.~30, 2017.

\bibitem{44}
A.~Dosovitskiy, L.~Beyer, A.~Kolesnikov, D.~Weissenborn, X.~Zhai,
  T.~Unterthiner, M.~Dehghani, M.~Minderer, G.~Heigold, S.~Gelly \emph{et~al.},
  ``An image is worth 16x16 words: Transformers for image recognition at
  scale,'' \emph{arXiv preprint arXiv:2010.11929}, 2020.

\bibitem{45}
N.~Carion, F.~Massa, G.~Synnaeve, N.~Usunier, A.~Kirillov, and S.~Zagoruyko,
  ``End-to-end object detection with transformers,'' in \emph{European
  conference on computer vision}.\hskip 1em plus 0.5em minus 0.4em\relax
  Springer, 2020, pp. 213--229.

\bibitem{46}
H.~Wang, Y.~Zhu, H.~Adam, A.~Yuille, and L.-C. Chen, ``Max-deeplab: End-to-end
  panoptic segmentation with mask transformers,'' in \emph{Proceedings of the
  IEEE/CVF conference on computer vision and pattern recognition}, 2021, pp.
  5463--5474.

\bibitem{47}
L.~Zhou, Y.~Zhou, J.~J. Corso, R.~Socher, and C.~Xiong, ``End-to-end dense
  video captioning with masked transformer,'' in \emph{Proceedings of the IEEE
  conference on computer vision and pattern recognition}, 2018, pp. 8739--8748.

\bibitem{50}
D.~Erhan, A.~Courville, Y.~Bengio, and P.~Vincent, ``Why does unsupervised
  pre-training help deep learning?'' in \emph{Proceedings of the thirteenth
  international conference on artificial intelligence and statistics}.\hskip
  1em plus 0.5em minus 0.4em\relax JMLR Workshop and Conference Proceedings,
  2010, pp. 201--208.

\bibitem{51}
C.~Gan, T.~Yao, K.~Yang, Y.~Yang, and T.~Mei, ``You lead, we exceed: Labor-free
  video concept learning by jointly exploiting web videos and images,'' in
  \emph{Proceedings of the IEEE Conference on Computer Vision and Pattern
  Recognition}, 2016, pp. 923--932.

\bibitem{52}
A.~Owens and A.~A. Efros, ``Audio-visual scene analysis with self-supervised
  multisensory features,'' in \emph{Proceedings of the European Conference on
  Computer Vision (ECCV)}, 2018, pp. 631--648.

\bibitem{53}
C.~Gan, B.~Gong, K.~Liu, H.~Su, and L.~J. Guibas, ``Geometry guided
  convolutional neural networks for self-supervised video representation
  learning,'' in \emph{Proceedings of the IEEE conference on computer vision
  and pattern recognition}, 2018, pp. 5589--5597.

\bibitem{54}
Z.~Wu, Y.~Xiong, S.~X. Yu, and D.~Lin, ``Unsupervised feature learning via
  non-parametric instance discrimination,'' in \emph{Proceedings of the IEEE
  conference on computer vision and pattern recognition}, 2018, pp. 3733--3742.

\bibitem{58}
C.~Wei, L.~Xie, X.~Ren, Y.~Xia, C.~Su, J.~Liu, Q.~Tian, and A.~L. Yuille,
  ``Iterative reorganization with weak spatial constraints: Solving arbitrary
  jigsaw puzzles for unsupervised representation learning,'' in
  \emph{Proceedings of the IEEE/CVF Conference on Computer Vision and Pattern
  Recognition}, 2019, pp. 1910--1919.

\bibitem{59}
T.~Chen, S.~Kornblith, M.~Norouzi, and G.~Hinton, ``A simple framework for
  contrastive learning of visual representations,'' in \emph{International
  conference on machine learning}.\hskip 1em plus 0.5em minus 0.4em\relax PMLR,
  2020, pp. 1597--1607.

\bibitem{57}
B.~Wu, W.-H. Cheng, Y.~Zhang, Q.~Huang, J.~Li, and T.~Mei, ``Sequential
  prediction of social media popularity with deep temporal context networks,''
  \emph{arXiv preprint arXiv:1712.04443}, 2017.

\bibitem{60}
H.-Y. Lee, J.-B. Huang, M.~Singh, and M.-H. Yang, ``Unsupervised representation
  learning by sorting sequences,'' in \emph{Proceedings of the IEEE
  international conference on computer vision}, 2017, pp. 667--676.

\bibitem{61}
H.~Cho, T.~Kim, H.~J. Chang, and W.~Hwang, ``Self-supervised spatio-temporal
  representation learning using variable playback speed prediction,''
  \emph{arXiv preprint arXiv:2003.02692}, vol.~2, pp. 13--14, 2020.

\bibitem{77}
L.~Lin, S.~Song, W.~Yang, and J.~Liu, ``Ms2l: Multi-task self-supervised
  learning for skeleton based action recognition,'' in \emph{Proceedings of the
  28th ACM International Conference on Multimedia}, 2020, pp. 2490--2498.

\bibitem{75}
N.~Zheng, J.~Wen, R.~Liu, L.~Long, J.~Dai, and Z.~Gong, ``Unsupervised
  representation learning with long-term dynamics for skeleton based action
  recognition,'' in \emph{Proceedings of the AAAI Conference on Artificial
  Intelligence}, vol.~32, no.~1, 2018.

\bibitem{76}
K.~Su, X.~Liu, and E.~Shlizerman, ``Predict \& cluster: Unsupervised skeleton
  based action recognition,'' in \emph{Proceedings of the IEEE/CVF Conference
  on Computer Vision and Pattern Recognition}, 2020, pp. 9631--9640.

\bibitem{78}
S.~Xu, H.~Rao, X.~Hu, J.~Cheng, and B.~Hu, ``Prototypical contrast and reverse
  prediction: Unsupervised skeleton based action recognition,'' \emph{IEEE
  Transactions on Multimedia}, 2021.

\bibitem{79}
H.~Rao, S.~Xu, X.~Hu, J.~Cheng, and B.~Hu, ``Augmented skeleton based
  contrastive action learning with momentum lstm for unsupervised action
  recognition,'' \emph{Information Sciences}, vol. 569, pp. 90--109, 2021.

\bibitem{80}
L.~Li, M.~Wang, B.~Ni, H.~Wang, J.~Yang, and W.~Zhang, ``3d human action
  representation learning via cross-view consistency pursuit,'' in
  \emph{Proceedings of the IEEE/CVF Conference on Computer Vision and Pattern
  Recognition}, 2021, pp. 4741--4750.

\bibitem{70}
P.~Wang, J.~Wen, C.~Si, Y.~Qian, and L.~Wang, ``Contrast-reconstruction
  representation learning for self-supervised skeleton-based action
  recognition,'' \emph{arXiv preprint arXiv:2111.11051}, 2021.

\bibitem{73}
T.~Guo, H.~Liu, Z.~Chen, M.~Liu, T.~Wang, and R.~Ding, ``Contrastive learning
  from extremely augmented skeleton sequences for self-supervised action
  recognition,'' in \emph{Proceedings of the AAAI Conference on Artificial
  Intelligence}, vol.~36, no.~1, 2022, pp. 762--770.

\bibitem{55}
K.~Hassani and A.~H. Khasahmadi, ``Contrastive multi-view representation
  learning on graphs,'' in \emph{International Conference on Machine
  Learning}.\hskip 1em plus 0.5em minus 0.4em\relax PMLR, 2020, pp. 4116--4126.

\bibitem{56}
J.~Qiu, Q.~Chen, Y.~Dong, J.~Zhang, H.~Yang, M.~Ding, K.~Wang, and J.~Tang,
  ``Gcc: Graph contrastive coding for graph neural network pre-training,'' in
  \emph{Proceedings of the 26th ACM SIGKDD International Conference on
  Knowledge Discovery \& Data Mining}, 2020, pp. 1150--1160.

\bibitem{64}
L.~N. Smith, ``Cyclical focal loss,'' \emph{arXiv preprint arXiv:2202.08978},
  2022.

\bibitem{63}
L.~N. Smith, ``General cyclical training of neural networks,'' \emph{arXiv
  preprint arXiv:2202.08835}, 2022.

\bibitem{65}
A.~Shahroudy, J.~Liu, T.-T. Ng, and G.~Wang, ``Ntu rgb+ d: A large scale
  dataset for 3d human activity analysis,'' in \emph{Proceedings of the IEEE
  conference on computer vision and pattern recognition}, 2016, pp. 1010--1019.

\bibitem{66}
J.~Liu, A.~Shahroudy, M.~Perez, G.~Wang, L.-Y. Duan, and A.~C. Kot, ``Ntu rgb+
  d 120: A large-scale benchmark for 3d human activity understanding,''
  \emph{IEEE transactions on pattern analysis and machine intelligence},
  vol.~42, no.~10, pp. 2684--2701, 2019.

\bibitem{81}
J.~Wang, X.~Nie, Y.~Xia, Y.~Wu, and S.-C. Zhu, ``Cross-view action modeling,
  learning and recognition,'' in \emph{Proceedings of the IEEE conference on
  computer vision and pattern recognition}, 2014, pp. 2649--2656.

\bibitem{83}
O.~Moliner, S.~Huang, and K.~{\AA}str{\"o}m, ``Bootstrapped representation
  learning for skeleton-based action recognition,'' in \emph{Proceedings of the
  IEEE/CVF Conference on Computer Vision and Pattern Recognition}, 2022, pp.
  4154--4164.

\bibitem{84}
F.~M. Thoker, H.~Doughty, and C.~G. Snoek, ``Skeleton-contrastive 3d action
  representation learning,'' in \emph{Proceedings of the 29th ACM International
  Conference on Multimedia}, 2021, pp. 1655--1663.

\bibitem{86}
L.~Van~der Maaten and G.~Hinton, ``Visualizing data using t-sne.''
  \emph{Journal of machine learning research}, vol.~9, no.~11, 2008.

\end{thebibliography}

\ifCLASSOPTIONcaptionsoff
\newpage
\fi
\end{document}